\documentclass{article}

\usepackage[final]{neurips_2022}

\usepackage[utf8]{inputenc} 
\usepackage[T1]{fontenc}    
\usepackage{hyperref}       
\usepackage{url}            
\usepackage{booktabs}       
\usepackage{amsfonts}       
\usepackage{nicefrac}       
\usepackage{microtype}      
\usepackage{xcolor}         
\usepackage{multirow}
\usepackage{mathtools}
\usepackage{amssymb}
\usepackage{subcaption}

\usepackage{dsfont}
\usepackage{enumitem}

\usepackage{soul}
\usepackage{subcaption}
\usepackage{multirow}
\usepackage{booktabs}
\usepackage{csquotes}
\usepackage{graphicx}
\usepackage{mathtools}

\newtheorem{theorem}{\textbf{Theorem}}

\newtheorem{corollary}{\textbf{Corollary}}

\newtheorem{proposition}{\textbf{Proposition}}

\newcommand{\EE}{\mathbb{E}}

\newcommand{\bX}{\mathbf{X}}

\newcommand{\bZ}{\mathbf{Z}}

\newcommand{\bb}{\mathbf{b}}

\newcommand{\bx}{\mathbf{x}}
\newcommand{\by}{\mathbf{y}}
\newcommand{\bR}{\mathbf{R}}

\newcommand{\bp}{\mathbf{p}}

\newcommand{\ba}{\mathbf{a}}
\newcommand{\bv}{\mathbf{v}}

\newcommand{\bu}{\mathbf{u}}

\newcommand{\bA}{\mathbf{A}}
\newcommand{\bB}{\mathbf{B}}
\newcommand{\bC}{\mathbf{C}}

\newcommand{\bS}{\mathbf{S}}

\newcommand{\bE}{\mathbf{E}}
\newcommand{\bL}{\mathbf{L}}

\newcommand{\bI}{\mathbf{I}}

\newcommand{\bz}{\mathbf{z}}

\newcommand{\be}{\mathbf{e}}

\newcommand{\bD}{\mathbf{D}}

\newcommand{\bW}{\mathbf{W}}

\newcommand{\sm}{\mathbf{A}}

\newcommand{\eps}{\epsilon}

\newcommand{\certradius}{R}
\newcommand{\manifold}{\mathcal{M}}
\newcommand{\seta}{\mathcal{A}}
\newcommand{\setb}{\mathcal{B}}

\def\archname{\mathrm{LipConvnet}}
\def\blockname{\mathrm{LipBlock}}
\def\lln{\mathrm{LLN}}
\def\crclip{$\mathrm{CRC}$-$\mathrm{Lip}$}

\def\fastsoc{Fast}
\def\pooling{Pool}
\def\cert{CRC}

\usepackage{enumitem}

\setlist[description]{leftmargin=*,labelindent=*}
\setlist[itemize]{leftmargin=*}
\setlist[enumerate]{leftmargin=*}

\title{Improved techniques for deterministic $l_2$ robustness}

\author{%
   Sahil Singla \\ \\
   Department of Computer Science\\
   University of Maryland\\
   \texttt{ssingla@umd.edu} \\ 
   \And
   Soheil Feizi \\ \\
   Department of Computer Science\\
   University of Maryland\\
   \texttt{sfeizi@umd.edu} \\ 
}

\begin{document}

\maketitle

\begin{abstract}
Training convolutional neural networks (CNNs) with a strict $1$-Lipschitz constraint under the $l_{2}$ norm is useful for adversarial robustness, interpretable gradients and stable training. $1$-Lipschitz CNNs are usually designed by enforcing each layer to have an orthogonal Jacobian matrix (for all inputs) to prevent the gradients from vanishing during backpropagation. However, their performance often significantly lags behind that of heuristic methods to enforce Lipschitz constraints where the resulting CNN is not \textit{provably} $1$-Lipschitz. In this work, we reduce this gap by introducing (a) a procedure to certify robustness of $1$-Lipschitz CNNs by replacing the last linear layer with a $1$-hidden layer MLP that significantly improves their performance for both standard and provably robust accuracy, (b) a method to significantly reduce the training time per epoch for Skew Orthogonal Convolution (SOC) layers ($>30\%$ reduction for deeper networks) and (c) a class of pooling layers using the mathematical property that the $l_{2}$ distance of an input to a manifold is $1$-Lipschitz. Using these methods, we significantly advance the state-of-the-art for standard and provable robust accuracies on CIFAR-10 (gains of  $+1.79\%$ and $+3.82\%$) and similarly on CIFAR-100 (+$3.78\%$ and +$4.75\%$) across all networks. Code is available at \url{https://github.com/singlasahil14/improved_l2_robustness}. 

\end{abstract}






\section{Introduction}
The Lipschitz constant \footnote{In this work, we assume the Lipschitz constant under the $l_{2}$ norm.} of a neural network $f: \mathbb{R}^{d} \to \mathbb{R}^{c}$, denoted by $\mathrm{Lip}(f)$, controls the change in the output divided by change in the input (both changes measured in the $l_{2}$ norm). Previous work provides evidence that a small Lipschitz constant is useful for interpretable saliency maps \citep{tsipras2019robustness}, generalization bounds \citep{long2019sizefree}, Wasserstein distance estimation \citep{villani2012optimal}, adversarial robustness \citep{42503} and preventing gradient explosion during backpropagation \citep{xiao2018dynamical}. Several prior works \citep{miyato2018spectral, gulrajani2017improved} use heuristic methods to enforce Lipschitz constraints to successfully address problems such as stabilizing GAN training. However, these methods do not enforce a guaranteed lipschitz bound and it remains challenging to achieve strong results with provable guarantees.



Using the composition property i.e. $\mathrm{Lip}(g\ \circ\ h) \leq \mathrm{Lip}(g)\  \mathrm{Lip}(h)$, we can construct a $1$-Lipschitz neural network by constraining each layer to be $1$-Lipschitz. However, a key difficulty with this approach is that because a $1$-Lipschitz layer can only reduce the gradient norm during backpropagation, for deeper networks, this results in small gradient values for layers closer to the input making training slow and difficult. To address this problem, \citet{Anil2018SortingOL} introduce Gradient Norm Preserving (GNP) architectures where each layer preserves the gradient norm during backpropagation. For $1$-Lipschitz Convolutional Neural Networks (CNNs), this involves using orthogonal convolutions (convolution layers with an orthogonal Jacobian matrix) \citep{li2019lconvnet, trockman2021orthogonalizing, singlafeizi2021, yu2022constructing, projunn2021} and using a class of GNP activation functions called $\mathrm{HouseHolder}$ activations \citep{singla2022improved}. 

\begin{table}
\centering
\caption{CIFAR-10 results using faster gradients, projection pooling, CRC-Lip}
\begin{tabular}{l | ll| l | ll | ll}
\toprule
SOC  &  \multicolumn{2}{|c|}{Time/epoch (secs)} &  Reduction & \multicolumn{2}{c|}{Standard accuracy} & \multicolumn{2}{|c}{Provable robust accuracy}\\
layers & Ours & Previous &  & Ours & Previous & Ours & Previous\\
\midrule
6 & \textbf{25.34} & 30.63 & \textbf{-17.27\%} & \textbf{79.36\%} & 76.68\% & \textbf{67.13\%} & 60.09\%  \\
11 & \textbf{37.11} & 48.44 & \textbf{-23.39\%} & \textbf{79.57\%} & 77.73\% & \textbf{66.75\%} & 62.82\% \\
16 & \textbf{49.27} & 66.74 & \textbf{-26.17\%} & \textbf{79.44\%} & 77.78\% & \textbf{66.99\%} & 62.75\% \\
21 & \textbf{61.59} & 83.18 & \textbf{-25.96\%} & \textbf{79.13\%} & 77.50\% & \textbf{66.45\%} & 63.31\% \\
26 & \textbf{71.51} & 100.70 & \textbf{-28.99\%} & \textbf{79.19\%} & 77.18\% & \textbf{66.28\%} & 62.46\% \\
31 &  \textbf{84.00} & 119.55 & \textbf{-29.74\%} & \textbf{78.64\%} & 74.43\% & \textbf{66.05\%} & 59.65\% \\
36 & \textbf{95.06} & 137.10 & \textbf{-30.66\%} & \textbf{78.57\%} & 72.73\% & \textbf{65.94\%} & 57.18\% \\
41 & \textbf{106.01} & 156.25 & \textbf{-32.16\%} & \textbf{78.41\%} & 71.33\% & \textbf{65.51\%} & 55.74\% \\
\bottomrule
\end{tabular}
\label{table:fast_grad}
\end{table}

Among orthogonal convolutions, SOC \citep{singlafeizi2021} achieves state-of-the-art results on CIFAR-10, 100 \citep{singla2022improved}. SOC uses the following two mathematical properties to construct an orthogonal convolution: (a) the exponential of a skew symmetric matrix is orthogonal and (b) the matrix exponential can be computed using the exponential series. A drawback of using the exponential series is that it requires us to apply the convolution operation multiple times to achieve a reasonable approximation of an orthogonal matrix. Consequently, during backpropagation, computing the gradient with respect to the weights for a single SOC layer requires us to compute the weight gradient for each of these convolutions resulting in significant training overhead. In this work, we show that because the matrix is skew symmetric, using some approximations, the gradients with respect to weights can be computed in significantly reduced time with no loss in performance. This enables us to reduce the training time per epoch using SOC by $>30\%$ for networks with large number of SOC layers (Results in Table \ref{table:fast_grad}). 


Another limitation of $1$-Lipschitz CNNs is that their performance is often significantly below compared to that of standard CNNs. Recently, \cite{singla2022improved} introduced a procedure for certifying robustness by relaxing the orthogonality requirements of the last linear layer (i.e. the linear layer mapping penultimate neurons to class logits) achieving state of the art results on CIFAR-10, 100 \citep{Krizhevsky09learningmultiple}. Since MLPs are more expressive than linear layers, one would expect improved \textit{standard accuracy} by replacing the last linear layer with them. However, the resulting networks are not $1$-Lipschitz and achieving high robustness guarantees (\textit{provable robust accuracy}) is difficult.  


To certify robustness for these networks, note that since the mapping from input to penultimate layer is $1$-Lipschitz, the robustness certificate for the penultimate output also provides a certificate for the input. Thus, we first replace the linear layer mapping penultimate output to logits with a $1$-hidden layer MLP (Multi layer Perceptron) because such MLPs are easier to certify compared to deep MLPs. To certify robustness for the MLP, we use the Curvature-based Robustness Certificate or CRC \citep{2020curvaturebased}. To train MLP to have high robustness guarantees, we use a variant of adversarial training that only applies adversarial perturbations to the MLP (not the whole network). We call our certification procedure $\mathrm{CRC}$-$\mathrm{Lip}$. This results in improved results for both the standard and also the provable robust accuracy  across all network architectures on CIFAR-10 ($ \geq 1.63\%,\ \geq 3.14\%$) and CIFAR-100 ($\geq 2.49\%,\ \geq 2.27\%$ respectively). Results are in Tables \ref{table:results_cifar10} and \ref{table:results_cifar100}.



While several works have attempted to construct novel and more expressive orthogonal convolution layers and GNP activation functions, current state-of-the-art $1$-Lipschitz CNNs still use pooling layers by taking the $\max$ of different elements. In this work, we introduce a class of $1$-Lipschitz pooling layers called \textit{projection pooling} using the following mathematical property: Given a manifold $\manifold \subset \mathbb{R}^{n}$ and input $\bx \in \mathbb{R}^{n}$, the function $d_{\manifold} : \mathbb{R}^{n} \to \mathbb{R}$ defined as the $l_2$ distance of $\bx$ to $\manifold$ is $1$-Lipschitz. Thus, to construct a pooling layer, we can first define a \textit{learnable manifold} with parameters $\Theta$ (Example in Section \ref{sec:lip_pool}). During training, for input $\bx \in \mathbb{R}^{n}$, the pooling layer simply outputs $d_{\manifold}(\bx) \in \mathbb{R}$ as the output, resulting in decrease in input dimension by a factor of $n$. Moreover, since the output $d_{\manifold}(\bx)$ is also a function of $\Theta$, $\Theta$ can be learned during training. However, solving for the distance $d_{\manifold}(\bx)$ can be difficult especially when $\manifold$ is a high-dimensional manifold. 

To address this limitation, (a) we use $2\mathrm{D}$ projection pooling layers that reduce the dimension by factor of $2$ and (b) we construct these layers using piecewise linear curves for which the distance can be computed efficiently by computing the minimum distance to all line segments and the connecting points (Example in Appendix Figure \ref{fig:lip_pool}). If the curve is closed and without self-intersections, we can also define a signed projection pooling for which the signs of the output for points inside and outside the region enclosed by the curve are different ($\bx$ and $\by$ in Figure \ref{fig:lip_pool}). This allows the subsequent layers to distinguish between the inputs inside and outside the region. In this work, we show some preliminary results using a simple $2\mathrm{D}$ projection pooling layer (Section \ref{sec:lip_pool}). We leave the problem of constructing high performance $2\mathrm{D}$ projection pooling layers open for future research. 

In summary, in this paper, we make the following contributions:
\begin{itemize}
    \item We introduce a method for faster computation of the weight gradient for SOC layers. For deeper networks, we observe reduction in training time per epoch by $> 30\%$ (Table \ref{table:fast_grad}).   
    \item We introduce a certification procedure called $\mathrm{CRC}$-$\mathrm{Lip}$ that replaces the last linear layer with a $1$-hidden layer MLP and results in significantly improved standard and provable robust accuracy. For deeper networks ($\geq 35$ layers), we observe improvements of $\geq 5.84\%, \geq 8.00\%$ in standard and $\geq 8.76\%, \geq 8.65\%$ in provable robust accuracy ($l_{2}$ radius $36/255$) on CIFAR-10,100 respectively. 
    \item We introduce a large class of $1$-Lipschitz pooling layers called {\it projection pooling} using the mathematical property that the $l_{2}$ distance $d_{\manifold}(\bx)$ of an input $\bx$ to the manifold $\manifold$ is $1$-Lipschitz. 
    \item On CIFAR-10, across all architectures, we achieve the best standard and provable robust accuracy (at $36/255$) of $79.57, 67.13\%$ respectively (gain of $+1.79\%, +3.82\%$ from prior works). Similarly, on CIFAR-$100$, we achieve $51.84, 39.27\%$  ($+3.78\%, +4.75\%$ from prior works). These results establish new state-of-the-art results in the standard and provable robust accuracy on both datasets. 
    
\end{itemize}

\section{Related work}\label{sec:related_work}

\textbf{Provable defenses against adversarial examples}: For a provably robust classifier, we can guarantee that its predictions remain constant within some region around the input. Most of the existing methods for provable robustness either bound the Lipschitz constant or use convex relaxations \citep{polyhedradomain2017, NEURIPS2018_f2f44698,  NEURIPS2019_0a9fdbb1, abstractdomain2019,singh2018robustness, weng2018CertifiedRobustness, NEURIPS2019_246a3c55, zhang2018recurjac, zhang2018crown, Wong2018ScalingPA, Wong2017ProvableDA, Singh2018FastAE, Raghunathan2018SemidefiniteRF, Dvijotham18, croce2019provable, kulis2009kernelized, dj2020, Lu2020Neural, 2020curvaturebased, Bunel2020BranchAB, leino21gloro, leino2021relaxing, linfty2021, zhang2022boosting, wang2021betacrown, huang2021training, MLSYS2021_ca46c1b9, singh2021overcoming, palma2021scaling}. However, these methods are often not scalable to large neural networks while achieving high performance. In contrast, randomized smoothing \citep{Liu2018TowardsRN, Cao2017MitigatingEA, Lcuyer2018CertifiedRT, Li2018CertifiedAR, Cohen2019CertifiedAR, Salman2019ProvablyRD, levine2019certifiably, confidence_cert2020, cursedimensionalitykumar20, salman2020, certpatchsalman2021} scales to large neural networks but is a \textit{probabilistically certified defense}: certifying robustness with high probability requires generating a large number of noisy samples leading to high inference-time. The defense we propose in this work is deterministic and not comparable to randomized smoothing. 

\textbf{Provably Lipschitz neural networks}:
The class of Gradient Norm Preserving (GNP) and 1-Lipschitz fully connected neural networks was first introduced by \citet{Anil2018SortingOL}. To design each layer to be GNP, they orthogonalize weight matrices and use a class of piecewise linear GNP activations called $\mathrm{GroupSort}$. Later, \citet{singla2022improved} proved that for any piecewise linear GNP function to be continuous, different Jacobian matrices in a neighborhood must change via householder transformations, implying that $\mathrm{GroupSort}$ is a special case of more general $\mathrm{HouseHolder}$ activations. Several previous works enforce Lipschitz constraints on convolution layers using spectral normalization, clipping or regularization \citep{cisseparseval2017, Tsuzuku2018LipschitzMarginTS, qian2018lnonexpansive, gouk2020regularisation, sedghi2018singular}. However, these methods either enforce loose lipschitz bounds or do not scale to large networks. To ensure that the Lipschitz constraint on convolutional layers is tight, recent works construct convolution layers with an orthogonal Jacobian \citep{li2019lconvnet, trockman2021orthogonalizing, singlafeizi2021, yu2022constructing, paraunitary2021, projunn2021}. These approaches avoid the aforementioned issues and allow training of large, provably $1$-Lipschitz CNNs achieving state-of-the-art results for deterministic $l_{2}$ robustness. 

\section{Problem setup and Notation}\label{sec:problem_setup} 
For a vector $\bv$, $\bv_{j}$ denotes its $j^{th}$ element. For a matrix $\bA$, $\bA_{j,:}$ and $\bA_{:,k}$ denote the $j^{th}$ row and $k^{th}$ column respectively. Both $\bA_{j,:}$ and $\bA_{:,k}$ are assumed to be column vectors (thus $\bA_{j,:}$ is the transpose of $j^{th}$ row of $\bA$). $\bA_{j,k}$ denotes the element in $j^{th}$ row and $k^{th}$ column of $\bA$. 
$\bA_{:j,:k}$ denotes the matrix containing the first $j$ rows and $k$ columns of $\bA$. We define $\bA_{:j} = \bA_{:j, :}$ and $\bA_{j:} = \bA_{j:, :}$. Similar notation applies to higher order tensors. $\bI$ denotes the identity matrix, $\mathbb{R}$ to denote the field of real numbers. We construct a $1$-Lipschitz neural network, $f: \mathbb{R}^{d} \to \mathbb{R}^{c}$ ($d$ is the input dimension, $c$ is the number of classes) by composing $1$-Lipschitz convolution layers and GNP activation functions. We often use the abbreviation $f_{i} - f_{j}: \mathbb{R}^{d} \to \mathbb{R}$ to denote the function so that: 
$$\left(f_{i} - f_{j}\right)(\bx) = f_{i}(\bx) - f_{j}(\bx),\qquad \forall\ \bx \in \mathbb{R}^{d} $$
For a matrix $\bA \in \mathbb{R}^{q \times r}$ and a tensor $\bB \in \mathbb{R}^{p \times q \times r}$, $\overrightarrow{\bA}$ denotes the vector constructed by stacking the rows of $\bA$ and $ \overrightarrow{\bB}$ by stacking the vectors $\overrightarrow{\bB_{j,:,:}},\ j \in [p-1]$ so that: 
\begin{align*}
&\left(\overrightarrow{\bA}\right)^{T} = \begin{bmatrix}
\bA_{0,:}^{T}\ ,\ \bA_{1,:}^{T}\ ,\ \hdots\ ,\ \bA_{q-1,:}^{T}\end{bmatrix} ,\quad \left(\overrightarrow{\bB}\right)^{T} = \begin{bmatrix}
\left(\overrightarrow{\bB_{0,:,:}}\right)^{T}\ ,\ \left(\overrightarrow{\bB_{1,:,:}}\right)^{T}\ ,\ \hdots\ ,\ \left(\overrightarrow{\bB_{p-1,:,:}}\right)^{T} \end{bmatrix}
\end{align*}

For a $2$D convolution filter, $\bL \in \mathbb{R}^{p \times q \times r \times s}$ and input $\bX \in \mathbb{R}^{q \times n \times n}$, we use $\bL \star \bX \in \mathbb{R}^{p \times n \times n}$ to denote the convolution of filter $\bL$ with $\bX$. We use the the notation $\bL \star^{i} \bX \triangleq \bL \star^{i-1}\left(\bL \star \bX\right)$ and $\bL \star^{0} \bX = \bX$. Unless specified, we assume zero padding and stride 1 in each direction.

\section{Faster gradient computation for Skew Orthogonal Convolutions}\label{sec:fast_soc} 
Among the existing orthogonal convolution layers in the literature, Skew Orthogonal Convolutions (or SOC) by \cite{singlafeizi2021} achieves state-of-the-art results on CIFAR-10,100 \citep{singla2022improved}. SOC first constructs a convolution filter $\bL$ whose Jacobian is skew-symmetric. This is followed by a convolution exponential \citep{Hoogeboom2020TheCE} operation. Since the exponential of a skew-symmetric matrix is orthogonal, the Jacobian of the resulting layer is an orthogonal matrix. 

However, a drawback of this procedure is that convolution exponential requires multiple convolution operations per SOC layer to achieve a reasonable approximation of the orthogonal Jacobian. Consequently, if we use $k$ convolution operations in the SOC layer during forward pass, we need to compute the gradient with respect to the weights $\bL$ (called \emph{convolution weight gradient}) per convolution operation ($k$ times) which can lead to slower training time especially when the number of SOC layers is large. To address this limitation, we show that even if we use $k$ convolutions in the forward pass of an SOC layer, the weight gradient for the layer can be computed using a \textit{single convolution weight gradient} during backpropagation leading to significant reduction in training time. 
For simplicity, let us first consider the case of an orthogonal fully connected layer (i.e. not convolution) with the same input and output size ($n$). Later, we will see that our analysis leads to improvements for orthogonal convolutional layers. Further, assume that the weights i.e. $\sm \in \mathbb{R}^{n \times n}$ are skew-symmetric i.e. $\sm = - \sm^{T}$ and given the input $\bx \in \mathbb{R}^{n}$, the output $\bz  \in \mathbb{R}^{n}$ is computed as follows: 
\begin{align}
\bz = \left(\sum_{i=0}^{k} \frac{\sm^{i}}{i!}\right) \bx + \bb,\qquad \quad \text{ where }\ \ \sm = -\sm^{T} \label{eq:z_def}
\end{align}

We approximate the exponential series: $\exp(\sm) = \sum_{i=0}^{\infty} \sm^{i}/i!$ using a finite number of terms ($k$). \\
\textbf{Forward pass} $(\bz)$: To compute $\bz$ during the forward pass, we use the following iterations:
\begin{align}
&\bu^{(i)} = \begin{cases}
  \bx & i=k-1 \\
  \bx + (\sm\bu^{(i+1)})/(i+1) & i \leq k-2 
\end{cases} \label{eq:ui_finite}
\end{align}
It can be shown that $\bz = \bu^{(0)}$ (Appendix \ref{sec:proof_iterations}). During backpropagation, given the gradient of loss $\ell$ w.r.t. layer output $\bz$ i.e. $\nabla_{\bz}\  \ell$, we want to compute $\nabla_{\bx}\  \ell$ (input gradient) and $\nabla_{\sm}\  \ell$ (weight gradient). 

\textbf{Input gradient} $(\nabla_{\bx}\  \ell)$: To compute $\nabla_{\bx}\  \ell$, observe that $\bz$ is a linear function of $\bx$. Thus, using the chain rule, skew-symmetricity ($\sm^{T} = -\sm$) and the property $\left(\sm^{i}\right)^T = \left(\sm^T\right)^{i}$, we have:
\begin{align*}
\nabla_{\bx}\  \ell = \left(\sum_{i=0}^{k} \frac{\sm^{i}}{i!}\right)^{T} \left(\nabla_{\bz}\  \ell\right) = \left(\sum_{i=0}^{\infty} \frac{(-\sm)^{i}}{i!}\right) \left(\nabla_{\bz}\  \ell\right) 
\end{align*}
We can again approximate the exponential series using the same number of terms as in the forward pass i.e. $k$. To compute the finite term approximation, we use the following iterations:
\begin{align}
&\bv^{(i)} = \begin{cases}
  \nabla_{\bz}\  \ell  & i=k-1 \\
  \nabla_{\bz}\  \ell - (\sm\bv^{(i+1)})/(i+1) & i \leq k-2 
\end{cases} \label{eq:vi_finite}
\end{align}
Similar to forward pass, it can be shown that $\nabla_{\bx}\  \ell = \bv^{(0)}$ (Appendix \ref{sec:proof_iterations}). 


\textbf{Weight gradient} $(\nabla_{\sm}\  \ell)$: We first derive the exact expression for $\nabla_{\sm}\  \ell$ in the Theorem below:
\begin{theorem}\label{thm:exact_grad} The gradient of the loss function $\ell$ w.r.t $\sm$ i.e. $\nabla_{\sm}\  \ell$ is given by:
\begin{align}
    &\nabla_{\sm}\ \ell = -\sum_{i = 1}^{k} \left(\left(\sm^{i-1}\bx\right)\left(\bv^{(i)}\right)^{T} - \bv^{(i)}\left(\sm^{i-1}\bx\right)^{T} \right) \label{eq:grad_exact_main}
\end{align}
where $\bv^{(i)}$ is defined as in equation \eqref{eq:vi_finite}.
\end{theorem}
Note that the first outer product i.e.  $\left(\sm^{i-1}\bx\right)\left(\bv^{(i)}\right)^{T}$ and the second i.e.  $\left(\bv^{(i)}\right)\left(\sm^{i-1}\bx\right)^{T}$ are transpose of each other implying that each term in the summation is skew-symmetric. Although these outer products can be computed in a straightforward way for orthogonal \textit{fully connected} layers, this is not the case for orthogonal \textit{convolution} layers (SOC in this case). This is because, for a convolution filter $\bL \in \mathbb{R}^{p\times q \times r \times s}$, the term analogous to $\sm^{i-1}\bx$ is a patch of size $q \times r \times s$ and that analogous to $\bv^{(i)}$ is another patch of size $p \times 1 \times 1$ resulting in an outer product of the desired size $p\times q \times r \times s$ \emph{per patch}. Thus, for SOC layers, each term inside the summation is computed by summing over the outer products for all such patches. For large input sizes, the number of such patches can often be large, making this computation expensive. To address this limitation, we use the following approximation: 
\begin{align}
    &\nabla_{\sm}\ \ell \approx - \left(\bu^{(1)}\left(\bv^{(1)}\right)^{T} - \bv^{(1)}\left(\bu^{(1)}\right)^{T} \right) \label{eq:grad_approx_main}
\end{align}
In Appendix \ref{sec:reason_approximation}, we show that the above approximation is principled because after subtracting the exact and approximation gradients (equations \eqref{eq:grad_exact_main} and \eqref{eq:grad_approx_main}) and simplifying, each term in the resulting series is divided by a large value in its denominator ($\approx 0$). This approximation is useful because in equation \eqref{eq:grad_approx_main}, the outer product needs to be computed once whereas in equation \eqref{eq:grad_exact_main}, the outer products need to be computed $k$ times. Also, both $\bu^{(1)}$ and $\bv^{(1)}$ are computed while computing $\bu^{(0)} = \bz$ during the forward pass and $\bv^{(0)} = \nabla_{\bx}\ \bz$ during the backward pass using the recurrences in equations \eqref{eq:ui_finite} and \eqref{eq:vi_finite}. Thus, we can simply store $\bu^{(1)},\bv^{(1)}$ during the forward,backward pass respectively so that $\nabla_{\sm}\ \ell$ can be computed directly using equation \eqref{eq:grad_approx_main}. In our experiments, we observe that this leads to significant reduction in training time with almost no drop in performance (Tables \ref{table:fast_grad}, \ref{table:results_cifar10}, \ref{table:results_cifar100}). 

\section{Curvature-based Robustness Certificate}\label{sec:crc} 
A key property of $1$-Lipschitz CNNs is that the output of each layer is $1$-Lipschitz with respect to the input. Given an input $\bx \in \mathbb{R}^{d}$, consider the penultimate output $g(\bx)  \in \mathbb{R}^{m}$ and logits $f(\bx) \in \mathbb{R}^{c}$ for some $1$-Lipschitz CNN. Existing robustness certificates \citep{li2019lconvnet, singla2022improved} rely on the \emph{linearity} of the function from $g(\bx) \to f(\bx)$. However, since MLPs have higher expressive power than linear functions \citep{citeulike3561150}, one way to improve performance could be to replace this mapping with MLPs. However, certifying robustness for deep MLPs is difficult. 

To address these challenges, we first replace the mapping from $g(\bx) \to f(\bx)$ with a $1$-hidden layer MLP because they are easier to certify compared to deeper networks. Because computing exact certificates for ReLU networks is known to be NP-complete \citep{sinha2018certifiable}, we use the differentiable $\mathrm{Softplus}$ activation \citep{NIPS2000_44968aec} to certify robustness. To certify robustness, we use the Curvature-based Robustness Certificate or CRC \citep{2020curvaturebased} because it provides exact certificates for a significant fraction of inputs for shallow MLPs. We provide a brief review of CRC in Appendix Section \ref{sec:crc_review}. Let $g: \mathbb{R}^{d} \to \mathbb{R}^{m}$ be a $1$-Lipschitz continuous function and $h: \mathbb{R}^{m} \to \mathbb{R}^{c}$ be a $1$-hidden layer MLP such that $f = h \circ g$. Further, let $l$ be the predicted class for input $\bx$ i.e. $f_{l}(\bx) \geq \max_{i \neq l} f_{i}(\bx)$. Since $g$ is $1$-Lipschitz, it can be shown that if $h$ is provably robust in an $l_{2}$ radius $\certradius$ around input $g(\bx)$, then $f$ is also provably robust in the $l_{2}$ radius $\certradius$ around $\bx$. The resulting procedure is called $\mathrm{CRC}$-$\mathrm{Lip}$ and is given in the following proposition:

\begin{figure*}[t]
\centering
\begin{subfigure}{0.45\linewidth}
\centering
\includegraphics[trim=0cm 8cm 0cm 4cm, clip, width=\linewidth]{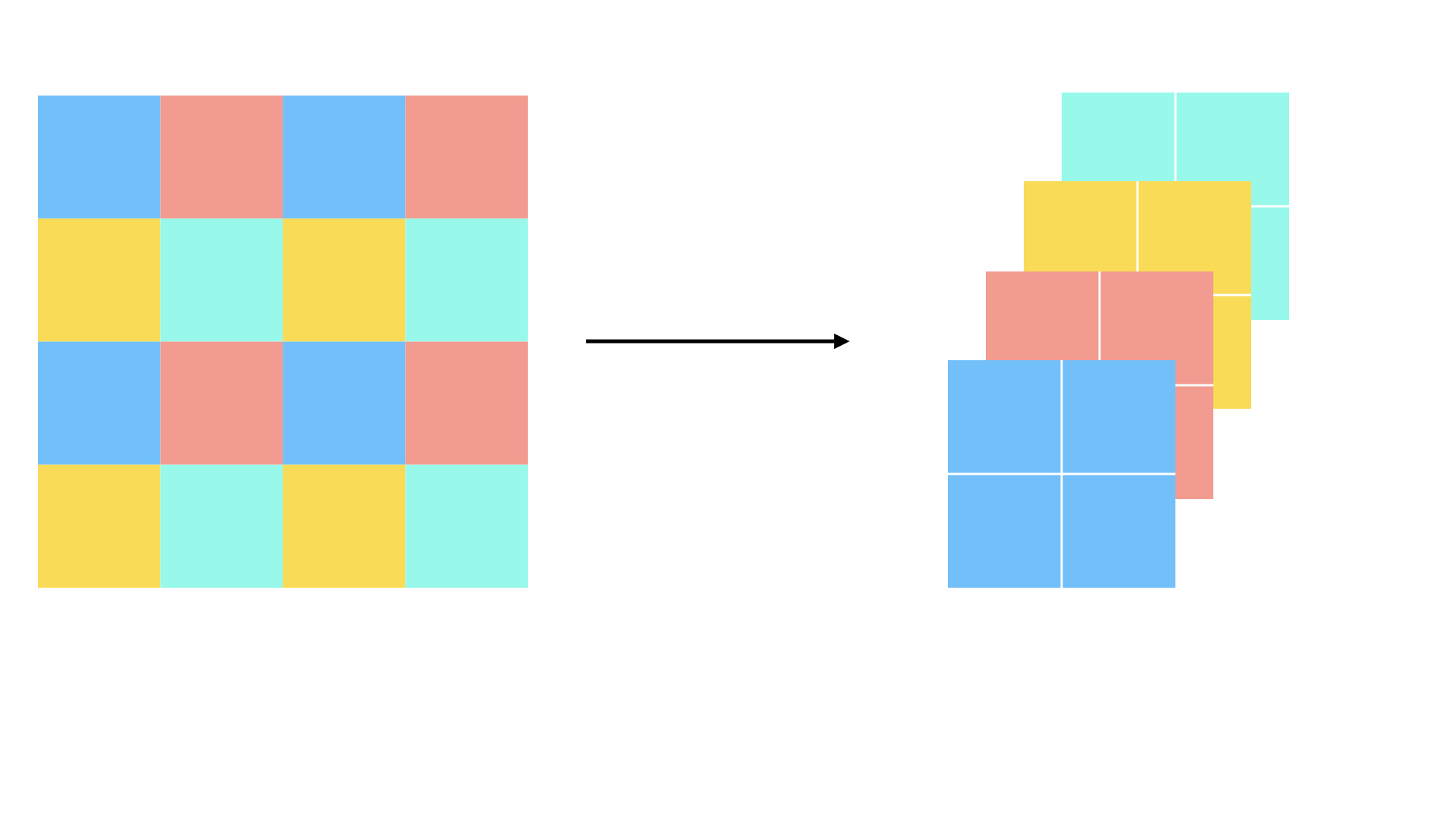}
\caption{Rearrangement operation  \citep{singlafeizi2021}}
\label{subfig:rearrange}
\end{subfigure}\quad
\begin{subfigure}{0.45\linewidth}
\centering
\includegraphics[trim=0cm 6cm 0cm 5cm, clip, width=\linewidth]{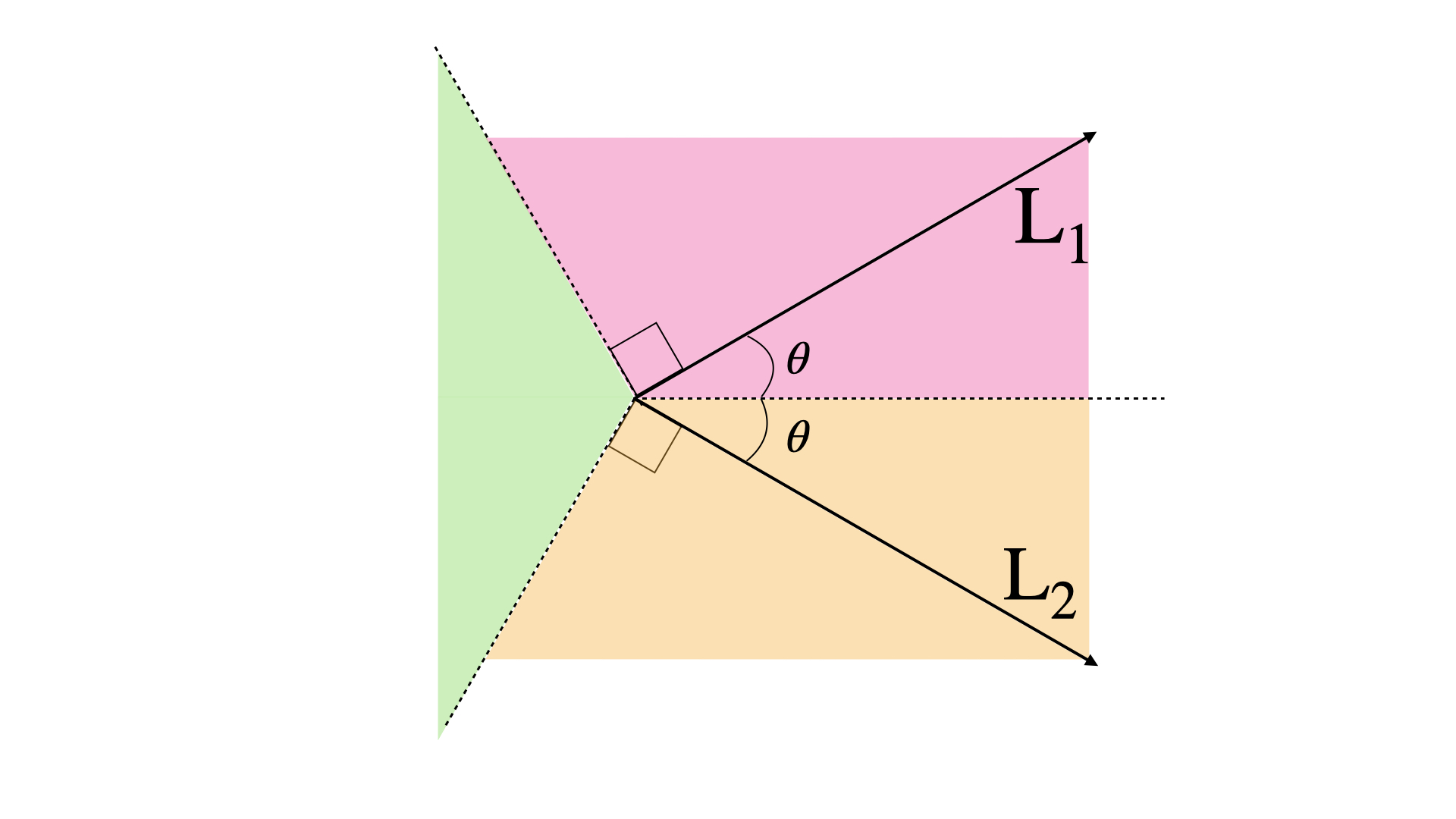}
\caption{2$\mathrm{D}$ projection pooling}
\label{subfig:projection_pool}
\end{subfigure}
\caption{Illustration of the rearrangement operation (left) and 2$\mathrm{D}$ projection pooling (right).}
\label{fig:hh_intro}
\end{figure*}



\begin{proposition}{$(\mathrm{CRC}$-$\mathrm{Lip})$}\label{prop:crc}
For input $\bx$ such that $f_{l}(\bx) \geq \max_{i \neq l} f_{i}(\bx)$, let $\certradius$ be the robustness certificate for $h$ using input $g(\bx)$, then $\certradius$ is also the robustness certificate for the function $f$:
\begin{align}
    \min_{i \neq l} \min_{\ h_{l}(\by^{*}) = h_{i}(\by^{*})} \|\by^{*} - g(\bx)\|_{2}\ \geq \certradius \implies \min_{i \neq l} \min_{\ f_{l}(\bx^{*}) = f_{i}(\bx^{*})} \|\bx^{*} - \bx\|_{2}\ \geq \certradius \label{eq:crc_cond}
\end{align}
\end{proposition}

Proof is in Appendix \ref{proof:crc}. In our experiments, we find that although replacing the linear layer with an MLP achieves high standard accuracy, directly using the above certificate often leads to very small robustness certificates and thus low certified robust accuracy. To address this problem, we introduce (a) an adversarial training procedure that \emph{only} applies to the input of the MLP $g(\bx)$ (i.e. not the input of the neural network $\bx$) and (b) curvature regularization to reduce the curvature of the MLP. It was shown in \citet{singlafeizi2021} that combining adversarial training with curvature regularization leads to significantly improved provable robust accuracy with small reduction in standard accuracy. This results in the following loss function for training:
\begin{align}
\min_{\Omega,\ \Psi}\quad \EE_{(\bx, l) \sim \mathcal{D}}\left[ \left(\max_{\|\by^{*} - g_{\Psi}(\bx)\|_{2} \leq \rho} \ell\left(h_{\Omega}(\by^{*}, l )\right)\right) + \gamma\ \mathcal{K}_{h} \right] \label{eq:loss_func}
\end{align}
In the above equation, $\Omega$ denote the parameters of the MLP (i.e. $h$), $\Psi$ are the parameters of the $1$-Lipschitz function $g$, $\mathcal{K}_{h}$ is the bound on the curvature of the MLP, $\rho$ denotes the $l_{2}$ perturbation radius and $\gamma$ denotes the curvature regularization coefficient. The curvature bound $\mathcal{K}_{h}$ is the same as in \citet{2020curvaturebased}. Also, since we apply adversarial perturbations in the penultimate layer, we also need to backpropagate through this procedure to enable training of previous layers. To this end, we simply use an identity map (same gradient output as the gradient input) to backpropagate through the adversarial training procedure and find that it works well in practice, achieving significantly better results compared to the state-of-the-art. It is possible that a more principled method of backpropagation may lead to better results and we leave that avenue open for future research.

\section{Projection pooling layers}\label{sec:lip_pool}
In $1$-Lipschitz CNNs, pooling is usually carried out as follows: given input $\bX \in \mathbb{R}^{q \times r \times r}$ ($r$ is even), we first use \emph{rearrangement} \citep{jacobsen2018irevnet} illustrated in Figure \ref{subfig:projection_pool} to construct $\bX' \in \mathbb{R}^{4q \times (r/2) \times (r/2)}$. Next, we apply an orthogonal convolution which gives an output of the same size $\bZ \in \mathbb{R}^{4q \times (r/2) \times (r/2)}$ and divide it into two tensors of equal sizes (along the channel dimension) giving $\bZ_{:2q} \in \mathbb{R}^{2q \times (r/2) \times (r/2)}$ and $\bZ_{2q:} \in \mathbb{R}^{2q \times (r/2) \times (r/2)}$. We then define $\max(\bZ_{:2q},\ \bZ_{2q:})$ (or either one of $\bZ_{:2q},\ \bZ_{2q:})$ as the output of the pooling layer. 

Although $\max$ is $1$-Lipschitz, its expressive power is limited. For example, consider the $1$-Lipschitz function $\|x, y\|_{2} = \sqrt{x^{2} + y^{2}}$. It is easy to see that if $x = y$, the error between $\|x, y\|_{2}$ and $\max(x, y)$ can be arbitrarily large for large values of $x, y$. To address such limitations, we construct expressive pooling layers using the following mathematical property: 
\begin{theorem}\label{thm:lip}
Given $\bx \in \mathbb{R}^{n}$ and manifold $\manifold \subset \mathbb{R}^{n}$, the distance function (in $l_2$ norm) is 1-Lipschitz:
\begin{align}
d_{\manifold}(\bx) = \min_{\bx^{*} \in \manifold}\|\bx^{*}-\bx\|_{2} \implies  \left|d_{\manifold}(\bx) - d_{\manifold}(\by)\right| \leq \|\bx - \by\|_{2} \label{eq:manifold_dist}
\end{align}
\end{theorem}
The above theorem provides a powerful method for constructing a \textit{learnable pooling layer} by selecting a \textit{learnable manifold} $\manifold_{\Theta} \subset \mathbb{R}^{r}$ ($\Theta$ denotes the set of learnable parameters). To apply the pooling operation to an input $\bx \in \mathbb{R}^{n}$, we simply output its $l_2$ distance to the manifold $\manifold_{\Theta}$ denoted by $d_{\manifold_{\Theta}}(\bx)$. Since $d_{\manifold_{\Theta}}(\bx) \in \mathbb{R}$ while $\bx \in \mathbb{R}^{n}$, this operation reduces the input size by a factor of $n$. 


As an example, let $\manifold_{\bu} = \{\bu\}$. Here $\bu \in \mathbb{R}^{n}$ is the learnable parameter and the distance function $d_{\manifold_{\bu}}(\bx) = \|\bx -\bu\|_{2}$. Even in this very simple case, for $\bu=\mathbf{0}$, $d_{\manifold_{\bu}}(\bx) = \|\bx\|_{2}$ and for $n=2$, this function can learn to represent the function $\|x,y\|_{2}$ discussed earlier exactly. We also prove that a \textit{signed} $l_2$ distance function can be defined when $\manifold$ satisfies certain properties: 

\begin{corollary}\label{cor:sign_lip}
If $\manifold$ is a connected manifold that divides $\mathbb{R}^{n}$ into two nonempty connected sets $\seta$ and $\setb$ such that $\seta \cap \setb = \phi$ and every path from $\ba \in \seta$ and $\bb \in \setb$ intersects a point on $\manifold$, then there exists a signed $l_{2}$ distance function with different signs in $\seta$ and $\setb$. 
\end{corollary}
Proofs of Theorems \ref{thm:lip} and Corollary \ref{cor:sign_lip} are given in Appendix \ref{proof:lip} and \ref{proof:sign_lip} respectively.

In practice, computing $d_{\manifold}$ can be difficult for high-dimensional $\manifold$. To tackle this issue, we use 2$\mathrm{D}$ pooling layers where $\manifold$ is defined to be a piecewise linear curve in 2$\mathrm{D}$. The $l_{2}$ distance can then be computed by finding the minimum distance to all the line segments and connecting points. We emphasize that even in this relatively simple case, $\manifold$ can have large number of parameters and $d_{\manifold}$ can still be efficient to compute because these individual distances can be computed in parallel. We use $\manifold_{\theta}$ defined below (illustrated in Figure \ref{subfig:projection_pool}, lines $L_{1},\ L_{2}$ correspond to $\phi = +\theta, -\theta$): 
\begin{align*}
\manifold_{\theta} = \left\{R(\cos \phi,\ \sin \phi):\quad R \geq 0,\ \phi \in \{+\theta, -\theta\} \right\}
\end{align*}
In each colored region (Figure \ref{subfig:projection_pool}), $d_{\manifold_{\theta}}$ can be computed using the following:
\begin{align}
d_{\manifold_{\theta}}\left(R(\cos \alpha,\ \sin \alpha)\right) = \begin{cases}
  R\sin(\alpha - \theta),  & 0 \leq \alpha \leq \theta + \pi/2 \\
  -R\sin(\alpha + \theta),  & 0 < -\alpha \leq \theta + \pi/2 \\
  R,  & \text{otherwise}
\end{cases}  \label{eq:dist_func}
\end{align}
To apply projection pooling on $\bZ_{:2q},\ \bZ_{2q:}$ discussed previously, we output
$d_{\manifold_{\theta}}\left(\bZ_{:2q}, \bZ_{2q:}\right)$.

\begin{table}[t]
\centering
\renewcommand{\arraystretch}{1.3} 
\begin{subtable}{.4\linewidth}
\centering
\begin{tabular}{ p{2cm}  p{2.5cm} }
\toprule
\textbf{Output Size} & \textbf{Layer} \\ 
\midrule
$32 \times 32 \times 32$ & $\mathrm{Conv}$  +  $\mathrm{MaxMin}$ \\
$64 \times 16 \times 16$ & $\mathrm{LipBlock}$-$\mathrm{n}/5$ \\
$128 \times 8 \times 8$ & $\mathrm{LipBlock}$-$\mathrm{n}/5$ \\
$256 \times 4 \times 4$ & $\mathrm{LipBlock}$-$\mathrm{n}/5$ \\
$512 \times 2 \times 2$ & $\mathrm{LipBlock}$-$\mathrm{n}/5$ \\
$1024 \times 1 \times 1$ & $\mathrm{LipBlock}$-$\mathrm{n}/5$ \\
\# of classes & $\mathrm{Linear / MLP}$-$1$\\
\bottomrule
\end{tabular}
\caption{$\mathrm{\archname}$-$\mathrm{n}$ Architecture}
\label{table:network_arch}
\end{subtable} \quad 
\begin{subtable}{.55\linewidth}
\centering
\begin{tabular}{ p{2.5cm} p{2.5cm}  p{1.25cm} }
\toprule
\textbf{Output Size} & \textbf{Layer} & \textbf{Repeats} \\ 
\midrule
$q \times r \times r$ & $\mathrm{Input}$ & \_ \\
\hline
$q \times r \times r$ & $\mathrm{Conv + MaxMin}$ & $\mathrm{n}/5-1$ \\
$4q \times (r/2) \times (r/2)$ & $\mathrm{Rearrange}$ & $1$ \\
$4q \times (r/2) \times (r/2)$ & $\mathrm{Conv}$ & $1$ \\
$2q \times (r/2) \times (r/2)$ & $\mathrm{Pooling}$ & $1$ \\
\bottomrule
\end{tabular}
\caption{$\blockname$-$\mathrm{n}/5$}
\label{table:block_arch}
\end{subtable}
\caption{$\archname$-$\mathrm{n}$ and $\blockname$-$\mathrm{n}/5$ architectures}
\end{table}

\section{Experiments}\label{sec:experiments} We perform experiments under the setting of provably robust image classification on CIFAR-10 and CIFAR-100 datasets. We use the $\archname$-5, 10, 15, $\dotsc$, 40 architectures for comparison. We use SOC as the orthogonal convolution and $\mathrm{MaxMin}$ as the activation in all architectures. All experiments were performed using $1$ NVIDIA GeForce RTX 2080 Ti GPU. All networks were trained for 200 epochs with initial learning rate of 0.1, dropped by a factor of 0.1 after 100 and 150 epochs. For adversarial training with curvature regularization, we use $\rho=36/255\ (0.1411),\ \gamma=0.5$ for CIFAR-$10$ and $\rho=0.2,\ \gamma=0.75$ for CIFAR-$100$. We find that certifying robustness using \crclip\ is computationally expensive for CIFAR-100 due to large number of classes. To address this issue, we only use classes with top-$10$ logits (instead of all $100$ classes) for CIFAR-100 (Table \ref{table:results_cifar100}). Since \crclip\  requires us to solve a convex optimization, we consider a certificate to be valid if the input is correctly classified and gradient at the optimal solution is $\leq 10^{-6}$ ($0$ otherwise). All results are reported using the complete test sets of CIFAR-$10$ and CIFAR-$100$. We compare the provable robust accuracy using 3 different $l_{2}$ perturbation radii: $36/255,\ 72/255,\ 108/255$. 

\begin{table}
    \centering
    \renewcommand{\arraystretch}{1.3} 
    \caption{Provable robustness results on CIFAR-10 (LipConvnet-$5, 10$ results in Appendix Table \ref{table:results_cifar10_appendix})}
    \label{table:results_cifar10}
    \centering
    \begin{tabular}{ll | llll | ll}
    \toprule
    \textbf{LipConv} & \textbf{Methods} & \multirow{2}{1.5cm}{\textbf{Standard Accuracy}} & \multicolumn{3}{c}{\textbf{Provable Robust Accuracy}} & \multicolumn{2}{|c}{\textbf{Increase}} \\ 
    \textbf{net}- &  &   &  36/255 &  72/255 &  108/255 & (standard) & (36/255) \\
\midrule
$15$ & Baseline &    77.78\% &  62.75\% &  46.34\% &   31.38\% &     \_ & \_ \\
              & \bf{+ \fastsoc} &    77.75\% &  62.52\% &  46.23\% &   31.19\% &    -0.03\% & -0.23\% \\
              & \bf{+ \cert} &    \textbf{79.44\%} &  \textbf{66.99\%} &  \textbf{52.56\%} &   \textbf{38.30\%} &    
              \textbf{+1.66\%} &
              \textbf{+4.24\%} \\
$20$ & Baseline &    77.50\% &  63.31\% &  46.42\% &   31.53\% & \_ & \_ \\
              & \bf{+ \fastsoc} &    77.13\% &  62.05\% &  45.86\% &   31.13\% &    -0.37\% & -1.26\% \\
              & \bf{+ \cert} &    \textbf{79.13\%} &  \textbf{66.45\%} &  \textbf{52.45\%} &   \textbf{38.12\%} &     \textbf{+1.63\%} & \textbf{+3.14\%} \\
$25$ & Baseline &    77.18\% &  62.46\% &  45.78\% &   31.16\% & \_ & \_ \\
              & \bf{+ \fastsoc} &    76.94\% &  61.91\% &  45.59\% &   30.69\% & -0.24\% & -0.55\% \\
              & \bf{+ \cert} &    \textbf{79.19\%} &  \textbf{66.28\%} &  \textbf{51.74\%} &   \textbf{37.99\%} &    \textbf{+2.01\%} &
              \textbf{+3.82\%} \\
$30$ & Baseline &    74.43\% &  59.65\% &  43.76\% &   29.16\% &     \_ & \_ \\
              & \bf{+ \fastsoc} &    74.69\% &  58.84\% &  43.33\% &   28.93\% &    +0.26\% & -0.81\% \\
              & \bf{+ \cert} &    \textbf{78.64\%} &  \textbf{66.05\%} &  \textbf{51.31\%} &   \textbf{37.30\%} &     \textbf{+4.21\%} & \textbf{+6.40\%} \\
$35$ & Baseline &    72.73\% &  57.18\% &  42.08\% &   28.09\% &     \_ & \_ \\
              & \bf{+ \fastsoc} &    72.91\% &  57.58\% &  41.52\% &   27.37\% &     +0.18\% & +0.40\% \\
              & \bf{+ \cert} &    \textbf{78.57\%} &  \textbf{65.94\%} &  \textbf{52.04\%} &   \textbf{37.63\%} &    \textbf{+5.84\%} & 
              \textbf{+8.76\%} \\
$40$ & Baseline &    71.33\% &  55.74\% &  39.32\% &   26.06\% & \_ & \_ \\
              & \bf{+ \fastsoc} &    71.60\% &  56.15\% &  39.82\% &   25.63\% &     +0.27\% & +0.41\% \\
              & \bf{+ \cert} &    \textbf{78.41\%} &  \textbf{65.51\%} &  \textbf{51.32\%} &   \textbf{37.30\%} &    \textbf{+7.08\%} & 
              \textbf{+9.77\%} \\
\bottomrule
\end{tabular}
\end{table}


\textbf{Table details:} For the baseline ("Baseline" in Tables \ref{table:results_cifar10}, \ref{table:results_cifar100}), we use the standard $\max$ pooling with the certificate based on $\lln$ \citep{singla2022improved} due to their superior performance over prior works. In Tables \ref{table:results_cifar10} and \ref{table:results_cifar100}, for each architecture, "\textbf{+ \fastsoc}" adds faster gradient computation, "\textbf{+ \cert}" replaces $\max$ pooling with projection pooling (equation \eqref{eq:dist_func}) and replaces the last linear layer with a $1$-hidden layer MLP (\crclip\ certificate) while also using faster gradients. For each architecture, the columns "Increase (Standard)" and "Increase (36/255)" denote the increase in standard and provable robust accuracy relative to "Baseline" standard and provable robust accuracy ($36/255$). Results where Projection pooling and \crclip\ are added separately are given in Appendix Tables \ref{table:results_cifar10_appendix} and \ref{table:results_cifar100_appendix}.

\textbf{LipConvnet Architecture:} We use a $1$-Lipschitz CNN architecture called $\archname$-$\mathrm{n}$ where $\mathrm{n}$ is a multiple of $5$ and $\mathrm{n} + 1$ is the total number of convolution layers. It consists of an initial SOC layer that expands the number of channels from $3$ to $32$. This is followed by $5$ blocks that reduce spatial dimensions (height and width) by half while doubling the number of channels. The architecture is summarized in Tables \ref{table:network_arch} and \ref{table:block_arch}. The last layer outputs the class logits and is either a linear layer in which case we use $\lln$ to certify robustness or a single-hidden layer MLP where we use \crclip .  



\textbf{Correcting the certificates}: Since SOC is an approximation, the Lipschitz constant of each SOC layer can be slightly more than $1$ and if we use a large number of SOC layers (e.g. $41$ in LipConvnet-$40$), the Lipschitz constant of the full network $(\mathrm{Lip}(f))$ can be significantly larger than $1$. To mitigate this issue, we take the following steps: (a) We use a large number of terms ($k=15$) during test time to approximate the exponential which results in a small approximation error (using the error bound in \citet{singlafeizi2021}) and (b) We compute $\mathrm{Lip}(f)$ by multiplying the lipschitz constant of all SOC layers (using the power method) and then divide the certificate by $\mathrm{Lip}(f)$.

\begin{table}
    \centering
    \renewcommand{\arraystretch}{1.3} 
    \caption{Provable robustness results on CIFAR-100 (LipConvnet-$5, 10$ results in Appendix Table \ref{table:results_cifar100_appendix})}
    \label{table:results_cifar100}
    \centering
    \begin{tabular}{ll | llll | ll}
    \toprule
    \textbf{LipConv} & \textbf{Methods} & \multirow{2}{1.5cm}{\textbf{Standard Accuracy}} & \multicolumn{3}{c}{\textbf{Provable Robust Accuracy}} & \multicolumn{2}{|c}{\textbf{Increase}} \\ 
    \textbf{net}- &  &   &  36/255 &  72/255 &  108/255 & (standard) & (36/255) \\
\midrule
$15$ & Baseline &    48.06\% &  34.52\% &  23.08\% &   14.70\% &     \_ & \_ \\
              &    \bf{ + \fastsoc} &    47.97\% &  33.84\% &  22.66\% &   14.26\% &    -0.09\% & -0.68\% \\
              & \bf{ + \cert} &    \textbf{50.79\%} &  \textbf{37.50\%} &  \textbf{26.16\%} &   \textbf{17.27\%} &     \textbf{+2.73\%} & \textbf{+2.98\%} \\
$20$ & Baseline &    47.37\% &  33.99\% &  23.40\% &   14.69\% &     \_ & \_ \\
              &    \bf{ + \fastsoc} &    46.41\% &  33.07\% &  22.06\% &   14.00\% &    -0.96\% & -0.92\% \\
              & \bf{ + \cert} &    \textbf{51.84\%} &  \textbf{38.54\%} &  \textbf{27.32\%} &   \textbf{18.53\%} &    
              \textbf{+4.47\%} & 
              \textbf{+4.55\%} \\
$25$ & Baseline &    45.77\% &  32.08\% &  21.36\% &   13.64\% &     \_ & \_ \\
              &    \bf{ + \fastsoc} &    45.28\% &  31.67\% &  20.69\% &   13.26\% &    -0.49\% & -0.41\% \\
              & \bf{ + \cert} &    \textbf{51.59\%} &  \textbf{39.27\%} &  \textbf{27.94\%} &   \textbf{19.06\%} &    \textbf{+5.82\%} & \textbf{+7.19\%} \\
$30$ & Baseline &    46.39\% &  33.08\% &  22.02\% &   13.77\% &     \_ & \_ \\
              &    \bf{ + \fastsoc} &    45.86\% &  32.54\% &  21.18\% &   12.77\% &    -0.53\% & -0.54\% \\
              & \bf{ + \cert} &    \textbf{50.97\%} &  \textbf{38.77\%} &  \textbf{27.73\%} &   \textbf{19.28\%} &    \textbf{+4.58\%} & \textbf{+5.69\%} \\
$35$ & Baseline &    43.42\% &  30.36\% &  19.71\% &   12.66\% &     \_ & \_ \\
              &    \bf{ + \fastsoc} &    42.78\% &  29.88\% &  19.73\% &   12.52\% &    -0.64\% & -0.48\% \\
              & \bf{ + \cert} &    \textbf{51.42\%} &  \textbf{39.01\%} &  \textbf{28.94\%} &   \textbf{20.29\%} &    \textbf{+8.00\%} & \textbf{+8.65\%} \\
$40$ & Baseline &    41.72\% &  28.53\% &  18.37\% &   11.49\% &     \_ & \_ \\
              &    \bf{ + \fastsoc} &    42.07\% &  28.51\% &  18.86\% &   11.89\% &    +0.35\% & -0.02\% \\
              & \bf{ + \cert} &    \textbf{50.11\%} &  \textbf{38.69\%} &  \textbf{28.45\%} &   \textbf{20.05\%} &   \textbf{+8.39\%} & 
              \textbf{+10.16\%} \\
\bottomrule
\end{tabular}
\end{table}


\subsection{Results using Faster gradient computation}
We show the reduction in training time per epoch (in seconds) on CIFAR-10 in Table \ref{table:fast_grad} and CIFAR-100 in Appendix Table \ref{table:results_cifar100_timings}. In both Tables, we observe that for deeper networks ($\geq 25$ layers), the reduction in time per epoch is $\approx 30\%$.  The corresponding standard and provable robust accuracy numbers are given in Tables \ref{table:results_cifar10} (CIFAR-10) and \ref{table:results_cifar100} (CIFAR-100) in the row "\textbf{+ \fastsoc}". From the columns "Increase (Standard)" and "Increase (36/255)", we observe that the performance is similar to the baseline across all network architectures. For deeper networks: $\archname$-$35, 40$, we observe an increase in performance for CIFAR-$10$ and small decrease ($< 0.64\%$) for CIFAR-$100$. 

\subsection{Results using projection pooling and CRC-Lip}
We observe that using CRC and projection pooling (row "\textbf{+ CRC}") leads to significant improvements in performance across all $\archname$ architectures. On CIFAR-$10$, we observe significant improvements in both the standard ($ \geq 1.63\%$, column "Increase (standard)") and provable robust accuracy ($\geq 3.14\%$, column "Increase (36/255)") across all architectures. On CIFAR-100, we also observe significant improvements in the standard ($ \geq 2.49\%$) and provable robust accuracy ($\geq 2.27\%$). For deeper networks ($\archname$-$35, 40$), on CIFAR-$10$, we observe even more significant gains in the standard ($ \geq 5.84\%$) and provable robust accuracy ($\geq 8.76\%$). Similarly on CIFAR-$100$, we observe gains of $\geq 8.00\%$ (standard) and $\geq 8.65\%$ (provable robust). 

Our results establish a new state-of-the-art for both the standard and provable robust accuracy across all attack radii. In Table \ref{table:results_cifar10} (CIFAR-10), the best "Baseline" standard and provable robust accuracy values (at $36/255, 72/255, 108/255$) are $77.78\%$ and $63.31\%,\ 46.42\%,\  31.53\%$ respectively. The best "\textbf{+ CRC}" values are $79.57\%$ and $67.13\%,\  53.17\%,\ 38.60\%$. This results in improvements of $+1.79\%$ and $+3.82\%, +6.75\%, +7.07\%$ respectively. Similarly, in Table \ref{table:results_cifar100} (CIFAR-100), the best "Baseline" standard and provable robust values are $48.06\%$ and $34.52\%, 23.40\%, 14.70\%$. The best "\textbf{+ CRC}" values are $51.84\%\  (+3.78\%)$ and $39.27\%\ (+4.75\%),\ 27.94\%\  (+4.54\%),\ 20.29\%\ (+5.59\%)$. 

\section{Acknowledgements}
This project was supported in part by NSF CAREER AWARD 1942230, a grant from NIST 60NANB20D134, HR001119S0026 (GARD), ONR YIP award N00014-22-1-2271, Army Grant No. W911NF2120076 and the NSF award CCF2212458.

\bibliographystyle{abbrvnat}
\bibliography{neurips_2022}

\clearpage







\section*{Checklist}


\begin{enumerate}

\item For all authors...
\begin{enumerate}
  \item Do the main claims made in the abstract and introduction accurately reflect the paper's contributions and scope?
    \answerYes{}
  \item Did you describe the limitations of your work?
    \answerYes{} in Section \ref{sec:experiments}, we discuss how \crclip might lead to slower certification.
  \item Did you discuss any potential negative societal impacts of your work?
    \answerNA{} 
  \item Have you read the ethics review guidelines and ensured that your paper conforms to them?
    \answerYes{}
\end{enumerate}

\item If you are including theoretical results...
\begin{enumerate}
  \item Did you state the full set of assumptions of all theoretical results?
    \answerYes{}
        \item Did you include complete proofs of all theoretical results?
    \answerYes{}
\end{enumerate}

\item If you ran experiments...
\begin{enumerate}
  \item Did you include the code, data, and instructions needed to reproduce the main experimental results (either in the supplemental material or as a URL)?
    \answerYes{}
  \item Did you specify all the training details (e.g., data splits, hyperparameters, how they were chosen)?
    \answerYes{}
        \item Did you report error bars (e.g., with respect to the random seed after running experiments multiple times)?
    \answerNA{} because it is expensive to run.
        \item Did you include the total amount of compute and the type of resources used (e.g., type of GPUs, internal cluster, or cloud provider)?
    \answerYes{}
\end{enumerate}

\item If you are using existing assets (e.g., code, data, models) or curating/releasing new assets...
\begin{enumerate}
  \item If your work uses existing assets, did you cite the creators?
    \answerYes{}
  \item Did you mention the license of the assets?
    \answerNA{}
  \item Did you include any new assets either in the supplemental material or as a URL?
    \answerNA{}
  \item Did you discuss whether and how consent was obtained from people whose data you're using/curating?
    \answerNA{}
  \item Did you discuss whether the data you are using/curating contains personally identifiable information or offensive content?
    \answerNA{}
\end{enumerate}

\item If you used crowdsourcing or conducted research with human subjects...
\begin{enumerate}
  \item Did you include the full text of instructions given to participants and screenshots, if applicable?
    \answerNA{}
  \item Did you describe any potential participant risks, with links to Institutional Review Board (IRB) approvals, if applicable?
    \answerNA{}
  \item Did you include the estimated hourly wage paid to participants and the total amount spent on participant compensation?
    \answerNA{}
\end{enumerate}

\end{enumerate}


\clearpage

\appendix


{\Large \bf Appendix}

\section{Proofs}\label{sec:proofs} 

\subsection{Proof of Theorem \ref{thm:exact_grad}}\label{sec:soc_gradient} 
Since $\sm$ is skew symmetric, $\forall\ j,k,\ \  \sm_{j,k} = -\sm_{k,j}$. \\
Thus, we focus on computing $\nabla_{\sm_{j,k}}\ \ell$ for $j > k$. 

\subsubsection{Preliminaries}
Using equation \eqref{eq:z_def} in the main text, we can rewrite $\bz$ as follows: 
\begin{align*}
    \bz = \bb + \sum_{i=0}^{\infty} \frac{\bz^{(i)}}{i!},\quad\text{where}\ \ \bz^{(i)} = \sm^{i}\bx 
\end{align*}
Using the chain rule, $\nabla_{\sm_{j,k}}\ \ell$ can be computed as follows:
\begin{align*}
&\nabla_{\sm_{j,k}}\ \ell = \sum_{i=0}^{\infty} \left(\nabla_{\sm_{j,k}}\ \bz^{(i)}\right)^T\left(\nabla_{\bz^{(i)}}\ \ell\right), \qquad \text{ where }\quad \nabla_{\bz^{(i)}}\ \ell =\frac{\nabla_{\bz}\ \ell}{i!} \\
\implies & \nabla_{\sm_{j,k}}\ \ell = \sum_{i=0}^{\infty} \left(\nabla_{\sm_{j,k}}\ \bz^{(i)}\right)^T\left(\frac{\nabla_{\bz}\ \ell}{i!}\right) = \left(\sum_{i=0}^{\infty} \frac{\nabla_{\sm_{j,k}}\ \bz^{(i)}}{i!}\right)^T \nabla_{\bz}\ \ell
\end{align*}
Moreover, since $\bz^{(0)}$ is independent of $\bW_{j,k}$, we have: $\nabla_{\sm_{j,k}}\ \bz^{(0)} = 0$ implying:
\begin{align}
& \nabla_{\sm_{j,k}}\ \ell = \left(\sum_{i=1}^{\infty} \frac{\nabla_{\sm_{j,k}}\ \bz^{(i)}}{i!}\right)^T \nabla_{\bz}\ \ell \label{eq:grad_wjk_def}
\end{align}

To compute each $\nabla_{\sm_{j,k}} \bz^{(i)}$, consider a matrix $\bE$ defined below:
\begin{align}
\bE_{p,q} = \begin{cases}
  +1  & p = j,\ q = k \\
  -1  & p = k,\ q = j \\
  0  & otherwise
\end{cases} \label{eq:E_def}
\end{align}

Now, let $\be^{(j)}$ denote a vector that satisfies:
\begin{align}
(\be^{(j)})_{l} = \begin{cases}
  1  & l = j\\
  0 & otherwise
\end{cases} \label{eq:smalle_def}
\end{align}
Clearly, $\bE$ can be rewritten as follows:
\begin{align}
\bE = \left(\be^{(j)}(\be^{(k)})^{T} - \be^{(k)}(\be^{(j)})^{T} \right) \label{eq:smalle_largeE}
\end{align}

\subsubsection{Computing $\nabla_{\sm_{j,k}} \bz^{(i)}$}

Since $\bz^{(i)} = \sm^{i}\bx$, we can compute $\nabla_{\sm_{j,k}} \bz^{(i)}$ as follows:
\begin{align}
    \nabla_{\sm_{j,k}} \bz^{(i)} &= \lim_{\eps \to 0} \frac{(\sm + \eps \bE) ^{i}\bx - \sm^{i}\bx}{\eps} = \lim_{\eps \to 0} \frac{(\sm + \eps \bE) ^{i} - \sm^{i}}{\eps}\bx \nonumber \\
     &= \lim_{\eps \to 0} \frac{ \sm^{i} + \eps\sum_{l=0}^{i-1} \left(\sm^{l}\right)\bE\left(\sm^{i-l-1}\right) + \eps^{2}(\text{higher order terms}) - \sm^{i}}{\eps}\bx \nonumber \\
    & = \sum_{l=0}^{i-1} \left(\sm^{l}\right)\bE\left(\sm^{i-l-1}\right)\bx \label{eq:grad_series} 
\end{align}

\subsubsection{Rewriting $\nabla_{\sm_{j,k}}\ \ell$}\label{sec:w_grad_sumeq}
Using equations \eqref{eq:grad_wjk_def} and \eqref{eq:grad_series}, we can simplify $\nabla_{\sm_{j,k}}\  \ell$ as follows:
\begin{align*}
\nabla_{\sm_{j,k}}\ \ell &= \left(\sum_{i=1}^{\infty} \frac{\nabla_{\sm_{j,k}}\ \bz^{(i)}}{i!}\right)^T \nabla_{\bz}\ \ell \\
&= \left(\sum_{i=1}^{\infty} \sum_{l=0}^{i-1} \frac{1}{i!} \left(\sm^{l}\right)\bE\left(\sm^{i-l-1}\bx\right)\right)^T \nabla_{\bz}\ \ell \\
&= \left(\sum_{i=1}^{\infty} \left(\sum_{l=0}^{i-1} \frac{1}{i!} \left(\sm^{l}\right)\bE\left(\sm^{i-l-1}\bx\right)\right)^T\right) \nabla_{\bz}\ \ell \\
&= \left(\sum_{i=1}^{\infty} \sum_{l=0}^{i-1} \frac{1}{i!} \left(\sm^{i-l-1}\bx\right)^{T}\bE^{T} \left(\sm^{l}\right)^{T} \right) \nabla_{\bz}\ \ell \\
&= \left(\sum_{i=1}^{\infty} \sum_{l=0}^{i-1} \frac{1}{i!} \left(\sm^{i-l-1}\bx\right)^{T}\bE^{T} \left(\sm^{T}\right)^{l} \right) \nabla_{\bz}\ \ell \\
&= \left(\sum_{i=1}^{\infty} \sum_{l=0}^{i-1} \frac{1}{i!} \left(\sm^{i-l-1}\bx\right)^{T}\bE^{T} \left(-\sm\right)^{l} \right) \nabla_{\bz}\ \ell \\
&= \sum_{i \geq 1,\ l \leq i-1} \left(\frac{1}{i!} \left(\sm^{i-l-1}\bx\right)^{T}\bE^{T} \left(\left(-\sm\right)^{l}\nabla_{\bz}\ \ell \right) \right) \\
&= \sum_{l \geq 0,\ i \geq l+1} \left(\frac{1}{i!} \left(\sm^{i-l-1}\bx\right)^{T}\bE^{T} \left(\left(-\sm\right)^{l}\nabla_{\bz}\ \ell \right) \right) 
\end{align*}
Rewriting $m=i-l-1$, we get:
\begin{align*}
\nabla_{\sm_{j,k}}\ \ell &= \sum_{l \geq 0,\ m \geq 0} \left(\frac{1}{(m+l+1)!} \left(\sm^{m}\bx\right)^{T}\bE^{T} \left(\left(-\sm\right)^{l}\nabla_{\bz}\ \ell \right) \right) 
\end{align*}
Note that the above equation can also be rewritten in the following equivalent forms:
\begin{align}
\nabla_{\sm_{j,k}} &= \sum_{l = 0}^{\infty} \left(\sum_{m = 0}^{\infty} \frac{\sm^{m}\bx}{(m+l+1)!} \right)^{T}\bE^{T} \left(\left(-\sm\right)^{l}\nabla_{\bz}\ \ell \right)  \label{eq:fwd_grad} \\
\nabla_{\sm_{j,k}} &= \sum_{m = 0}^{\infty}  \left(\sm^{m}\bx\right)^{T} \bE^{T} \left(\sum_{l = 0}^{\infty}\frac{\left(-\sm\right)^{l}\nabla_{\bz}\ \ell}{(m+l+1)!}  \right)  \label{eq:bwd_grad}
\end{align}
Let us now define the following quantities:
\begin{align}
\bu^{(l)} &= \sum_{m = 0}^{\infty} \frac{\sm^{m}\bx}{(m+l)!} \label{eq:u_def_fwd} \\
\bv^{(m)} &= \sum_{l = 0}^{\infty}\frac{\left(-\sm\right)^{l}\nabla_{\bz}\ \ell}{(m+l)!} \label{eq:v_def_bwd}
\end{align}
Note that they are similar to equations \eqref{eq:ui_finite} and \eqref{eq:vi_finite} in the maintext. \\
Using equations \eqref{eq:u_def_fwd} and \eqref{eq:v_def_bwd}, we can rewrite \eqref{eq:fwd_grad} and \eqref{eq:bwd_grad} as follows:
\begin{align}
\nabla_{\sm_{j,k}} &= \sum_{l = 1}^{\infty} \left(\bu^{(l)}\right)^{T}\bE^{T} \left(\left(-\sm\right)^{l-1}\nabla_{\bz}\ \ell \right) \label{eq:fwd_grad_simple} \\
\nabla_{\sm_{j,k}} &= \sum_{m = 1}^{\infty} \left(\sm^{m-1}\bx\right)^{T} \bE^{T} \bv^{(m)}  \label{eq:bwd_grad_simple}
\end{align}
Using the definition of $\bE$ (equation \eqref{eq:E_def}), we obtain the following equivalent expressions of $\nabla_{\sm_{j,k}}\ \ell$:
\begin{align}
    &\nabla_{\sm_{j,k}}\ \ell = -\sum_{l = 1}^{\infty} \left(\bu^{(l)}_{j}\left(\left(-\sm\right)^{l-1}\nabla_{\bz}\ \ell \right)_{k} - \bu^{(l)}_{k}\left(\left(-\sm\right)^{l-1}\nabla_{\bz}\ \ell \right)_{j}\right) \label{eq:grad_scalar_exact_u} \\
    &\nabla_{\sm_{j,k}}\ \ell = -\sum_{m = 1}^{\infty} \left(\left(\sm^{m-1}\bx\right)_{j}\bv^{(m)}_{k} - \left(\sm^{m-1}\bx\right)_{k}\bv^{(m)}_{j}\right) \label{eq:grad_scalar_exact_v}
\end{align}
Using the above equation, we obtain the following equivalent expressions for $\nabla\ \ell$:
\begin{align}
    &\nabla_{\sm}\ \ell = -\sum_{l = 1}^{\infty} \left(\bu^{(l)}\left(\left(-\sm\right)^{l-1}\nabla_{\bz}\ \ell\right)^{T} - \left(\left(-\sm\right)^{l-1}\nabla_{\bz}\ \ell\right)\left(\bu^{(l)}\right)^{T}\right) \label{eq:grad_full_exact_u} \\
    &\nabla_{\sm}\ \ell = -\sum_{m = 1}^{\infty}  \left(\left(\sm^{m-1}\bx\right)\left(\bv^{(m)}\right)^{T} - \bv^{(m)}\left(\sm^{m-1}\bx\right)^{T} \right) \label{eq:grad_full_exact_v}
\end{align}

\subsection{Proof of Proposition \ref{prop:crc}}\label{proof:crc}
Consider the inequality given in equation \eqref{eq:crc_cond}:
\begin{align*}
    \min_{i \neq l} \min_{\ h_{l}(\by^{*}) = h_{i}(\by^{*})} \|\by^{*} - g(\bx)\|_{2}\ \geq \certradius 
\end{align*}
The above inequality says that for the prediction of $h$ to be different from $l$ for some input $\by^{*}$, the $l_{2}$ distance $\|\by^{*} - g(\bx)\|_{2}$ must be at least $\certradius$. \\
We first take the contrapositive of this statement:
\begin{align}
\|\by^{*} - g(\bx)\|_{2}\ \leq \certradius  \implies \forall_{i}\ h_{l}(\by^{*}) \geq h_{i}(\by^{*}) \label{eq:h_lip}
\end{align}
But since $g$ is $1$-Lipschitz, we have:
\begin{align}
\|\bx^{*} - \bx\|_{2} \leq \certradius \implies \|g(\bx^{*}) - g(\bx)\|_{2} \leq \certradius \label{eq:g_lip}
\end{align}
Using both \eqref{eq:h_lip} and \eqref{eq:g_lip}, we get: 
\begin{align*}
\|\bx^{*} - \bx\|_{2} \leq \certradius \implies \forall_{i}\ h_{l}(g(\bx^{*})) \geq h_{i}(g(\bx^{*})) \implies \forall_{i}\ f_{l}(\bx^{*}) \geq f_{i}(\bx^{*})
\end{align*}
Thus, in an $l_2$ ball of radius $\certradius$, prediction $l$ remains unchanged and the function is provably robust. 

\subsection{Proof of Theorem \ref{thm:lip}}\label{proof:lip}
Let $\bx^*,\ \by^* \in \manifold$ so that $d_{\manifold}(\bx)$ and $d_{\manifold}(\by)$ satisfy:
\begin{align}
d_{\manifold}(\bx) = \|\bx^{*}-\bx\|_{2} \qquad d_{\manifold}(\by) = \|\by^{*}-\by\|_{2} \label{eq:step0}
\end{align}
Since $\bx^{*}$ and $\by^{*}$ minimize the $l_2$ distance of $\bx$ and $\by$ respectively to $\manifold$, we have:
\begin{align}
&\|\bx^{*}-\bx\|_{2} \leq \|\by^{*}-\bx\|_{2}, \qquad \|\by^{*}-\by\|_{2} \leq \|\bx^{*}-\by\|_{2} \label{eq:step1}
\end{align}
Using the triangle inequality for the expressions $\|\by^{*}-\bx\|_{2}$ and $\|\bx^{*}-\by\|_{2}$, we have:
\begin{align}
&\|\by^{*}-\bx\|_{2} \leq \|\by^{*}-\by\|_{2} + \|\by-\bx\|_{2}, \qquad \|\bx^{*}-\by\|_{2} \leq \|\bx^{*}-\bx\|_{2} + \|\bx-\by\|_{2} \label{eq:step2}
\end{align}
Using inequalities \eqref{eq:step1} and \eqref{eq:step2}, we get:
\begin{align}
\|\bx^{*}-\bx\|_{2} \leq \|\by^{*}-\by\|_{2} + \|\by-\bx\|_{2}, \qquad \|\by^{*}-\by\|_{2} \leq \|\bx^{*}-\bx\|_{2} + \|\bx-\by\|_{2}  \label{eq:step3}
\end{align}
But then using equation \eqref{eq:step0} and inequality \eqref{eq:step3}, we get:
\begin{align*}
&d_{\manifold}(\bx) \leq d_{\manifold}(\by) + \|\by-\bx\|_{2}, \qquad d_{\manifold}(\by) \leq d_{\manifold}(\bx) + \|\bx-\by\|_{2} \\
&d_{\manifold}(\bx) - d_{\manifold}(\by) \leq \|\by-\bx\|_{2}, \qquad d_{\manifold}(\by) - d_{\manifold}(\bx) \leq \|\bx-\by\|_{2} \\
& \implies \left|d_{\manifold}(\bx) - d_{\manifold}(\by)\right| \leq \|\bx-\by\|_{2}
\end{align*}

\subsection{Proof of Corollary \ref{cor:sign_lip}}\label{proof:sign_lip}
We define a function $d_{\manifold}(\bx)$ that is (without loss of generality) positive in $\seta$ and negative in $\setb$:
\begin{align*}
d_{\manifold}(\bx) = \begin{cases}
  \bx \in \seta & +\min_{\bx^{*} \in \manifold}\|\bx^{*}-\bx\|_{2} \\
  \bx \in \setb & -\min_{\bx^{*} \in \manifold}\|\bx^{*}-\bx\|_{2} \\
\end{cases}
\end{align*}

Using Theorem \ref{thm:lip}, it is easy to see that within each set $\seta$ and $\setb$, $d_{\manifold}$ is $1$-Lipschitz and continuous. \\
Similarly, because every line segment connecting $\seta$ and $\setb$ crosses through some point on the $\manifold$, it is evident that $d_{\manifold}$ is continuous on each path connecting $\seta$ to $\setb$. \\
So, we next prove that $d_{\manifold}$ is $1$-Lipschitz if $\bx \in \seta$ and $\by \in \setb$ i.e.:
\begin{align*}
    \left|d_{\manifold}(\bx) - d_{\manifold}(\by)\right| \leq \|\bx - \by\|_{2} 
\end{align*}
Let $\bp \in \manifold$ be a point on the line segment connecting $\bx$ and $\by$ such that:
\begin{align}
& \|\bx - \by\|_{2} = \|\bx - \bp\|_{2} + \|\bp - \by\|_{2} \label{eq:step1_cor} 
\end{align}
But then by the definitions of $d_{\manifold}(\bx)$ and $d_{\manifold}(\by)$:
\begin{align}
& \left|d_{\manifold}(\bx)\right| \leq \|\bx - \bp\|_{2}, \qquad \left|d_{\manifold}(\by)\right| \leq \|\bp - \by\|_{2} \label{eq:step2_cor}
\end{align}
Using equation \eqref{eq:step1_cor} and inequality \eqref{eq:step2_cor}, we get:
\begin{align}
& \left|d_{\manifold}(\bx)\right| + \left|d_{\manifold}(\by)\right| \leq \|\bx - \bp\|_{2} + \|\bp - \by\|_{2} = \|\bx - \by\|_{2} \nonumber \\
& \implies \left|d_{\manifold}(\bx)\right| + \left|d_{\manifold}(\by)\right| \leq \|\bx - \by\|_{2} \label{eq:step3_cor}
\end{align}
Since $d_{\manifold}(\bx)$ and $d_{\manifold}(\by)$ have opposite signs, we have:
\begin{align}
    \left|d_{\manifold}(\bx)\right| + \left|d_{\manifold}(\by)\right| = \left|d_{\manifold}(\bx) - d_{\manifold}(\by)\right| \label{eq:step4_cor}
\end{align}
Using inequality \eqref{eq:step3_cor} and equation \eqref{eq:step4_cor}, we prove:
\begin{align*}
\left|d_{\manifold}(\bx) - d_{\manifold}(\by)\right| \leq \|\bx - \by\|_{2} 
\end{align*}
Hence, $\left|d_{\manifold}(\bx) - d_{\manifold}(\by)\right|$ is $1$-Lipschitz. 

\section{Approximating the convolution weight gradient}\label{sec:reason_approximation}

\subsection{Simplifying the expression for $\nabla_{\sm_{j,k}}\ \ell$}
Recall the expression for $\nabla_{\sm_{j,k}}\ \ell$ using equations \eqref{eq:grad_wjk_def} and \eqref{eq:grad_series}:
\begin{align*}
    & \nabla_{\sm_{j,k}}\ \ell = \left(\sum_{i=1}^{\infty} \frac{\nabla_{\sm_{j,k}}\ \bz^{(i)}}{i!}\right)^T \nabla_{\bz}\ \ell, \qquad \text{where}\ \ 
    \nabla_{\sm_{j,k}} \bz^{(i)} = \sum_{l=0}^{i-1} \left(\sm^{l}\right)\bE\left(\sm^{i-l-1}\right)\bx 
\end{align*}
We replace $\nabla_{\sm_{j,k}} \bz^{(i)}$ in $\nabla_{\sm_{j,k}}\  \ell$ and take $\bx$ out of the transpose:
\begin{align*}
\nabla_{\sm_{j,k}}\ \ell &= \bx^{T}\left(\sum_{i=1}^{\infty} \frac{1}{i!}\sum_{l=0}^{i-1} \left(\sm^{l}\right)\bE\left(\sm^{i-l-1}\right) \right)^T \nabla_{\bz}\ \ell \\ 
&= \bx^{T}\left(\bE + \frac{1}{2!}\left(\bE\sm + \sm\bE\right) + \frac{1}{3!}\left(\bE\sm^2 + \sm\bE\sm + \sm^2\bE\right) + \dotsc  \right)^{T} \nabla_{\bz}\ \ell  
\end{align*}
We simplify the above as follows:
\begin{align}
&\nabla_{\sm_{j,k}}\ \ell \label{eq:full_exact} \\
&= \bx^{T}\left(\bE + \frac{1}{2!}\left(\bE\sm + \sm\bE\right) + \frac{1}{3!}\left(\bE\sm^2 + \sm^2\bE\right) +
\frac{1}{4!}\left(\bE\sm^3 + \sm^3\bE\right) + \dotsc  \right)^{T} \nabla_{\bz}\ \ell \label{eq:part1_exact} \\
&\qquad + \bx^{T}\bR^{T} \left(\nabla_{\bz}\ \ell\right) \label{eq:part2_exact}
\end{align}
where $\bR$ is defined as follows:
\begin{align}
&\bR = \frac{1}{3!}\left(\sm\bE\sm\right) + \frac{1}{4!}\left( \sm\bE\sm^2 + \sm^2\bE\sm \right) + \frac{1}{5!}\left(\sm\bE\sm^{3} + \sm^{2}\bE\sm^{2} + \sm^{3}\bE\sm \right) + \dotsc \label{eq:R_def}
\end{align}
In the above simplification, $\bR$ contains all terms of the form $\sm^{l}\bE\sm^{m}$ where both $l \geq 1$ and $m \geq 1$.

\subsection{Simplifying the expression for $\left(\bu^{(1)}\right)^{T}\bE^{T}\bv^{(1)}$}
Recall that $\bu^{(1)}, \bv^{(1)}$ are given by (equations \eqref{eq:u_def_fwd} and \eqref{eq:v_def_bwd}):
\begin{align}
\bu^{(1)} = \left(\sum_{l = 1}^{\infty} \frac{\sm^{l-1}}{l!}\right)\bx, \qquad
\bv^{(1)} = \left(\sum_{m = 1}^{\infty}\frac{\left(-\sm\right)^{m-1}}{m!}\right) \nabla_{\bz}\ \ell \label{eq:ui_vi_full}
\end{align}
We first replace $\bv^{(1)}$ using the above in $\left(\bu^{(1)}\right)^{T}\bE^{T}\bv^{(1)}$:
\begin{align}
&\left(\bu^{(1)}\right)^{T}\bE^{T}\bv^{(1)} = \left(\bu^{(1)}\right)^{T}\bE^{T}\left(\sum_{m = 1}^{\infty}\frac{\left(-\sm\right)^{m-1}}{m!}\right) \nabla_{\bz}\ \ell \label{eq:transposed_eq}
\end{align}
Given $2$ matrices $\bB, \bC$, we have $\left(\bB^{T}\right)^{i} = \left(\bB^{i}\right)^{T}$ and $(\bB + \bC)^T = \bB^T + \bC^T$. Thus:
\begin{align}
\left(\sum_{m = 1}^{\infty}\frac{\sm^{m-1}}{m!}\right)^{T} = \left(\sum_{m = 1}^{\infty}\frac{\left(-\sm\right)^{m-1}}{m!}\right) \nonumber
\end{align}
Replacing the above in equation \eqref{eq:transposed_eq}, we get:
\begin{align*}
&\left(\bu^{(1)}\right)^{T}\bE^{T}\bv^{(1)} = \left(\bu^{(1)}\right)^{T}\bE^{T}\left(\sum_{m = 1}^{\infty}\frac{\sm^{m-1}}{m!}\right)^T \nabla_{\bz}\ \ell 
\end{align*}
Given $3$ matrices, $\bB, \bC, \bD$, we have $\bD^{T}\bC^{T}\bB^{T} = (\bB\bC\bD)^{T}$. Thus:
\begin{align*}
&\left(\bu^{(1)}\right)^{T}\bE^{T}\bv^{(1)} = \left(\left(\sum_{m = 1}^{\infty}\frac{\sm^{m-1}}{m!}\right)\bE\bu^{(1)}\right)^{T} \nabla_{\bz}\ \ell 
\end{align*}
Now, we replace $\bu^{(1)}$ using equation \eqref{eq:ui_vi_full} and take $\bx$ out of the transpose: 
\begin{align*}
&\left(\bu^{(1)}\right)^{T}\bE^{T}\bv^{(1)} = \bx^{T}\left(\left(\sum_{m = 1}^{\infty}\frac{\sm^{m-1}}{m!}\right)\bE\left(\sum_{l = 1}^{\infty} \frac{\sm^{l-1}}{l!}\right)\right)^{T} \nabla_{\bz}\ \ell 
\end{align*}
Expanding the two series, we get:
\begin{align*}
&\left(\bu^{(1)}\right)^{T}\bE^{T}\bv^{(1)} = \bx^{T}\left(\left(\bI + \frac{\sm}{2!} +  \frac{\sm^{2}}{3!} + \dotsc \right)\bE\left(\bI + \frac{\sm}{2!} +  \frac{\sm^{2}}{3!} + \dotsc\right)\right)^{T} \nabla_{\bz}\ \ell 
\end{align*}
We simplify the above as follows:
\begin{align}
&\left(\bu^{(1)}\right)^{T}\bE^{T}\bv^{(1)} \label{eq:full_approx}  \\
& = \bx^{T}\left(\bE + \frac{1}{2!}\left(\bE\sm + \sm\bE\right) + \frac{1}{3!}\left(\bE\sm^2 + \sm^2\bE\right) +
\frac{1}{4!}\left(\bE\sm^3 + \sm^3\bE\right) + \dotsc  \right)^{T} \nabla_{\bz}\ \ell \label{eq:part1_approx} \\
&\qquad \quad + \bx^{T}\bS^{T} \left(\nabla_{\bz}\ \ell\right) \label{eq:part2_approx}
\end{align}

where $\bS$ is defined as follows:
\begin{align}
&\bS = \left(\frac{\sm\bE\sm}{2!2!}\right) +  \left(\frac{\sm\bE\sm^2}{2!3!} + \frac{\sm^2\bE\sm}{3!2!}\right) + \left(\frac{\sm\bE\sm^3}{2!4!} + \frac{\sm^2\bE\sm^2}{3!3!} + \frac{\sm^3\bE\sm}{4!2!}\right) + \dotsc \label{eq:S_def}
\end{align}

\subsection{Comparing the two expressions}
Comparing equations \eqref{eq:full_exact} and \eqref{eq:full_approx}, the first part in the two are same. \\
The first part are the equations \eqref{eq:part1_exact} and \eqref{eq:part1_approx} respectively.\\
The two expressions only differ in the terms $\bx^{T}\bR^{T}\left(\nabla_{\bz}\ \ell\right)$ and $\bx^{T}\bS^{T}\left(\nabla_{\bz}\ \ell\right)$.\\
Note that in $\bR$ (equation \eqref{eq:R_def}), the term $\sm^{l}\bE\sm^{m}$ is multiplied by $1/(l+m+1)!$.\\
And in $\bS$ (equation \eqref{eq:S_def}), the term $\sm^{l}\bE\sm^{m}$ is multiplied by $1/(l+1)!(m+1)!$.\\
Thus, since the quantities $\bR$ and $\bS$ are multiplied by factorial in the denominator.\\ 
For higher values of $l$ or $m$, both the expressions converge to $0$. \\
Moreover, $\bS - \bR$ for smaller values of $l, m$ can be directly computed using equations \eqref{eq:R_def} and \eqref{eq:S_def}:
\begin{align}
&\bS - \bR = \left(\frac{\sm\bE\sm}{12}\right) +  \left(\frac{\sm\bE\sm^2}{24} + \frac{\sm^2\bE\sm}{24}\right) + \left(\frac{\sm\bE\sm^3}{80} + \frac{\sm^2\bE\sm^2}{360/7} + \frac{\sm^3\bE\sm}{80}\right) + \dotsc \label{eq:RS_diff}
\end{align}
We can see that the denominator in the above terms are already very small even for $l+m=4$. \\
Thus, we simply assume $\bS-\bR \approx \mathbf{0}$.\\
This results in the following approximate expression for $\nabla_{\sm_{j,k}}\ \ell$:
\begin{align}
    \nabla_{\sm_{j,k}}\ \ell \approx \left(\bu^{(1)}\right)^{T}\bE^{T}\bv^{(1)}
\end{align}
Using the definition of $\bE$ (equation \eqref{eq:E_def}), we obtain the following expression of $\nabla_{\sm_{j,k}}\ \ell$:
\begin{align}
    \nabla_{\sm_{j,k}}\ \ell \approx -(\bu^{(1)}_{j}\bv^{(1)}_{k} - \bu^{(1)}_{k}\bv^{(1)}_{j}) \label{eq:grad_scalar}
\end{align}

Using equation \eqref{eq:grad_scalar}, the gradient with respect to matrix $\sm$ can be computed as follows:
\begin{align}
    \nabla_{\sm}\ \ell \approx -\left(\bu^{(1)}\left(\bv^{(1)}\right)^{T} - \bv^{(1)}\left(\bu^{(1)}\right)^{T}\right) \label{eq:grad_matrix}
\end{align}

\section{Iterations for $\bz$ and $\nabla_{\bz}\ \ell$}\label{sec:proof_iterations}
Recall that we use the following iterations to compute $\bz$ during the forward pass:
\begin{align*}
&\bu^{(i)} = \begin{cases}
  \bx & i=k-1 \\
  \bx + \left(\sm\bu^{(i+1)}\right)/(i+1) & i \leq k-2
\end{cases} 
\end{align*}
We have to prove that $\forall\ k,\ \bu^{(0)} = \bz$ where: 
\begin{align*}
    \bz = \sum_{i=0}^{k-1} \frac{\sm^{i}\bx}{i!}
\end{align*}
To prove the same, we use induction over $k$.\\
For $k=1$, $\bu^{(0)} = \bz = \bx$ by definition.\\
For $k \geq 2$, we assume that the statement is true for $j \leq k-1$ and try to prove for $k$. \\ 
For $k$, we have the following expressions for $\bu^{(k-1)}$ and $\bu^{(k-2)}$:
$$\bu^{(k-1)} = \bx,\quad \bu^{(k-2)} = \bx + \sm\bx/(k-1) $$
Next, we apply the iteration on $\bu^{(k-2)}$ till $\bu^{(0)}$.\\
By induction, the first term i.e. $\bx$ simply results in $\sum_{i=0}^{k-2} (\sm^i\bx)/i!$. \\
The second term i.e. $\sm\bx/k$ in $(\sm^{k-1}\bx)/(k-1)!$. Thus, we have:
$$\bu^{(0)} = \left(\sum_{i=0}^{k-2} \frac{\sm^{i}\bx}{i!}\right) + \frac{\sm^{k-1}\bx}{(k-1)!} = \sum_{i=0}^{k-1} \frac{\sm^{i}\bx}{i!}  $$
The same can proof follows for $\nabla_{\bz}\ \ell$ except that we replace $\sm$ with $-\sm$.

\section{Review of Curvature-based Robustness Certificate}\label{sec:crc_review}
Given a binary classifier $f: \mathbb{R}^{d} \to \mathbb{R}$ and input $\bx^{(0)}$, we are interested in finding its shortest $l_{2}$ distance to the decision boundary: $ \min_{\bx} \|\bx - \bx^{(0)}\|_{2} \text{ where }f(\bx)=0$. Using lagrange multiplier $\eta$ for the constraint $f(\bx)=0$, the \emph{primal} and \emph{dual} problems can be written as follows:
\begin{align*}
    \text{\emph{(primal)}} \quad \min_{\bx} \max_{\eta} \left[ \frac{1}{2}\|\bx - \bx^{(0)}\|^{2}_{2} + \eta f(\bx) \right] \geq \max_{\eta} \min_{\bx}  \left[ \frac{1}{2}\|\bx - \bx^{(0)}\|^{2}_{2} + \eta f(\bx) \right] \quad \text{\emph{(dual)}}
\end{align*}
\citet{2020curvaturebased} prove that: If $m\bI \preccurlyeq \nabla^{2}_{\bx} f \preccurlyeq M\bI$, the \emph{dual} minimization can be solved using convex optimization for $-1/M \leq \eta \leq -1/m$. \\
Furthermore, they show that if the solution $\bx$ of the solution satisfies $f(\bx) = 0$, the certificate is tight i.e. \emph{primal} = \emph{dual}.

\clearpage

\section{Complete results on CIFAR-10 and CIFAR-100}

\begin{table}[h!]
    \centering
    \renewcommand{\arraystretch}{1.3} 
    \caption{Results for provable robustness against adversarial examples on the CIFAR-10 dataset.}
    \label{table:results_cifar10_appendix}
    \centering
    \begin{tabular}{ll | llll | ll}
    \toprule
    \textbf{LipConv} & \textbf{Methods} & \multirow{2}{1.5cm}{\textbf{Standard Accuracy}} & \multicolumn{3}{c}{\textbf{Provable Robust Accuracy}} & \multicolumn{2}{|c}{\textbf{Increase}} \\ 
   \textbf{net-} &  &   &  36/255 &  72/255 &  108/255 & (standard) &(36/255) \\
\midrule
$5$ & Baseline &    76.68\% &  60.09\% &  42.89\% &   27.45\% & \_ & \_ \\
              & \bf{+ \fastsoc} &    76.03\% &  60.12\% &  42.90\% &   27.91\% & -0.65\% & +0.03\% \\
              & \bf{+ \pooling} &    77.53\% &  62.17\% &  44.94\% &   29.74\% & 
              +0.85\% & +2.08\% \\
              & \bf{+ \cert} &    \textbf{79.36\%} &  \textbf{67.13\%} &  \textbf{52.49\%} &   \textbf{38.19\%} & 
              \textbf{+2.68\%} & \textbf{+7.04\%} \\
$10$ & Baseline &    77.73\% &  62.82\% &  45.62\% &   30.09\% & \_ & \_ \\
              &  \bf{+ \fastsoc} &    77.76\% &  62.42\% &  45.48\% &   30.38\% &    +0.03\% & -0.40\% \\
              & \bf{+ \pooling} &    77.31\% &  62.02\% &  46.04\% &   30.99\% &    -0.42\% & -0.80\% \\
              & \bf{+ \cert} &    \textbf{79.57\%} &  \textbf{66.75\%} &  \textbf{53.17\%} &   \textbf{38.60\%} &     \textbf{+1.84\%} & \textbf{+3.93\%} \\
$15$ & Baseline &    77.78\% &  62.75\% &  46.34\% &   31.38\% &     \_ & \_ \\
              & \bf{+ \fastsoc} &    77.75\% &  62.52\% &  46.23\% &   31.19\% &    -0.03\% & -0.23\% \\
              & \bf{+ \pooling} &    77.75\% &  63.14\% &  47.01\% &   32.05\% &     -0.03\% & +0.39\% \\
              & \bf{+ \cert} &    \textbf{79.44\%} &  \textbf{66.99\%} &  \textbf{52.56\%} &   \textbf{38.30\%} &    
              \textbf{+1.66\%} &
              \textbf{+4.24\%} \\
$20$ & Baseline &    77.50\% &  63.31\% &  46.42\% &   31.53\% & \_ & \_ \\
              & \bf{+ \fastsoc} &    77.13\% &  62.05\% &  45.86\% &   31.13\% &    -0.37\% & -1.26\% \\
              & \bf{+ \pooling} &    77.61\% &  63.25\% &  47.10\% &   32.84\% &    +0.11\% & -0.06\% \\
              & \bf{+ \cert} &    \textbf{79.13\%} &  \textbf{66.45\%} &  \textbf{52.45\%} &   \textbf{38.12\%} &     \textbf{+1.63\%} & \textbf{+3.14\%} \\
$25$ & Baseline &    77.18\% &  62.46\% &  45.78\% &   31.16\% & \_ & \_ \\
              & \bf{+ \fastsoc} &    76.94\% &  61.91\% &  45.59\% &   30.69\% & -0.24\% & -0.55\% \\
              & \bf{+ \pooling} &    75.44\% &  60.13\% &  45.02\% &   30.44\% &    -1.74\% & -2.33\% \\
              & \bf{+ \cert} &    \textbf{79.19\%} &  \textbf{66.28\%} &  \textbf{51.74\%} &   \textbf{37.99\%} &    \textbf{+2.01\%} &
              \textbf{+3.82\%} \\
$30$ & Baseline &    74.43\% &  59.65\% &  43.76\% &   29.16\% &     \_ & \_ \\
              & \bf{+ \fastsoc} &    74.69\% &  58.84\% &  43.33\% &   28.93\% &    +0.26\% & -0.81\% \\
              & \bf{+ \pooling} &    74.12\% &  59.17\% &  42.76\% &   28.75\% &    -0.31\% & -0.48\% \\
              & \bf{+ \cert} &    \textbf{78.64\%} &  \textbf{66.05\%} &  \textbf{51.31\%} &   \textbf{37.30\%} &     \textbf{+4.21\%} & \textbf{+6.40\%} \\
$35$ & Baseline &    72.73\% &  57.18\% &  42.08\% &   28.09\% &     \_ & \_ \\
              & \bf{+ \fastsoc} &    72.91\% &  57.58\% &  41.52\% &   27.37\% &     +0.18\% & +0.40\% \\
              & \bf{+ \pooling} &    72.84\% &  56.71\% &  40.41\% &   26.62\% &    +0.11\% & -0.47\% \\
              & \bf{+ \cert} &    \textbf{78.57\%} &  \textbf{65.94\%} &  \textbf{52.04\%} &   \textbf{37.63\%} &    \textbf{+5.84\%} & 
              \textbf{+8.76\%} \\
$40$ & Baseline &    71.33\% &  55.74\% &  39.32\% &   26.06\% & \_ & \_ \\
              & \bf{+ \fastsoc} &    71.60\% &  56.15\% &  39.82\% &   25.63\% &     +0.27\% & +0.41\% \\
              & \bf{+ \pooling} &    70.19\% &  54.22\% &  37.51\% &   23.88\% &    -1.14\% & -1.52\% \\
              & \bf{+ \cert} &    \textbf{78.41\%} &  \textbf{65.51\%} &  \textbf{51.32\%} &   \textbf{37.30\%} &    \textbf{+7.08\%} & 
              \textbf{+9.77\%} \\
\bottomrule
\end{tabular}
\end{table}

\begin{table}
    \centering
    \renewcommand{\arraystretch}{1.3} 
    \caption{Results for provable robustness against adversarial examples on the CIFAR-100 dataset.}
    \label{table:results_cifar100_appendix}
    \centering
    \begin{tabular}{ll | llll | ll}
    \toprule
    \textbf{LipConv} & \textbf{Methods} & \multirow{2}{1.5cm}{\textbf{Standard Accuracy}} & \multicolumn{3}{c}{\textbf{Provable Robust Accuracy}} & \multicolumn{2}{|c}{\textbf{Increase}} \\ 
   \textbf{net-} &  &   &  36/255 &  72/255 &  108/255 & (standard) & (36/255) \\
\midrule
 $5$ & Baseline &    46.74\% &  32.45\% &  21.04\% &   12.88\% &     \_ & \_ \\
              &    \bf{ + \fastsoc} &    46.18\% &  32.15\% &  20.79\% &   12.87\% &    -0.56\% & -0.30\% \\
              & \bf{ + \pooling} &    47.47\% &  33.51\% &  22.06\% &   13.73\% &     +0.73\% & +1.06\% \\
              & \bf{ + \cert} &    \textbf{49.73\%} &  \textbf{36.05\%} &  \textbf{24.68\%} &   \textbf{16.31\%} &    \textbf{+2.99\%} & \textbf{+3.60\%} \\
$10$ & Baseline &    47.96\% &  34.33\% &  22.55\% &   14.23\% &     \_ & \_ \\
              &    \bf{ + \fastsoc} &    47.24\% &  33.79\% &  22.38\% &   14.36\% &    -0.72\% & -0.54\% \\
              & \bf{ + \pooling} &    47.32\% &  34.01\% &  22.28\% &   14.04\% &    -0.64\% & -0.32\% \\
              & \bf{ + \cert} &    \textbf{50.45\%} &  \textbf{36.60\%} &  \textbf{25.57\%} &   \textbf{16.93\%} &     \textbf{+2.49\%} & \textbf{+2.27\%} \\
$15$ & Baseline &    48.06\% &  34.52\% &  23.08\% &   14.70\% &     \_ & \_ \\
              &    \bf{ + \fastsoc} &    47.97\% &  33.84\% &  22.66\% &   14.26\% &    -0.09\% & -0.68\% \\
              & \bf{ + \pooling} &    47.44\% &  34.12\% &  22.59\% &   14.48\% &    -0.62\% & -0.40\% \\
              & \bf{ + \cert} &    \textbf{50.79\%} &  \textbf{37.50\%} &  \textbf{26.16\%} &   \textbf{17.27\%} &     \textbf{+2.73\%} & \textbf{+2.98\%} \\
$20$ & Baseline &    47.37\% &  33.99\% &  23.40\% &   14.69\% &     \_ & \_ \\
              &    \bf{ + \fastsoc} &    46.41\% &  33.07\% &  22.06\% &   14.00\% &    -0.96\% & -0.92\% \\
              & \bf{ + \pooling} &    46.54\% &  32.75\% &  21.64\% &   13.22\% &    -0.83\% & -1.24\% \\
              & \bf{ + \cert} &    \textbf{51.84\%} &  \textbf{38.54\%} &  \textbf{27.32\%} &   \textbf{18.53\%} &    
              \textbf{+4.47\%} & 
              \textbf{+4.55\%} \\
$25$ & Baseline &    45.77\% &  32.08\% &  21.36\% &   13.64\% &     \_ & \_ \\
              &    \bf{ + \fastsoc} &    45.28\% &  31.67\% &  20.69\% &   13.26\% &    -0.49\% & -0.41\% \\
              & \bf{ + \pooling} &    45.26\% &  31.87\% &  20.88\% &   13.65\% &    -0.51\% & -0.21\% \\
              & \bf{ + \cert} &    \textbf{51.59\%} &  \textbf{39.27\%} &  \textbf{27.94\%} &   \textbf{19.06\%} &    \textbf{+5.82\%} & \textbf{+7.19\%} \\
$30$ & Baseline &    46.39\% &  33.08\% &  22.02\% &   13.77\% &     \_ & \_ \\
              &    \bf{ + \fastsoc} &    45.86\% &  32.54\% &  21.18\% &   12.77\% &    -0.53\% & -0.54\% \\
              & \bf{ + \pooling} &    45.15\% &  31.53\% &  21.18\% &   13.14\% &    -1.24\% & -1.55\% \\
              & \bf{ + \cert} &    \textbf{50.97\%} &  \textbf{38.77\%} &  \textbf{27.73\%} &   \textbf{19.28\%} &    \textbf{+4.58\%} & \textbf{+5.69\%} \\
$35$ & Baseline &    43.42\% &  30.36\% &  19.71\% &   12.66\% &     \_ & \_ \\
              &    \bf{ + \fastsoc} &    42.78\% &  29.88\% &  19.73\% &   12.52\% &    -0.64\% & -0.48\% \\
              & \bf{ + \pooling} &    42.33\% &  28.65\% &  18.59\% &   11.75\% &    -1.09\% & -1.71\% \\
              & \bf{ + \cert} &    \textbf{51.42\%} &  \textbf{39.01\%} &  \textbf{28.94\%} &   \textbf{20.29\%} &    \textbf{+8.00\%} & \textbf{+8.65\%} \\
$40$ & Baseline &    41.72\% &  28.53\% &  18.37\% &   11.49\% &     \_ & \_ \\
              &    \bf{ + \fastsoc} &    42.07\% &  28.51\% &  18.86\% &   11.89\% &    +0.35\% & -0.02\% \\
              & \bf{ + \pooling} &    40.27\% &  27.09\% &  17.36\% &   10.62\% &    -1.45\% & -1.44\% \\
              & \bf{ + \cert} &    \textbf{50.11\%} &  \textbf{38.69\%} &  \textbf{28.45\%} &   \textbf{20.05\%} &   \textbf{+8.39\%} & 
              \textbf{+10.16\%} \\
\bottomrule
\end{tabular}
\end{table}

\clearpage

\section{Running time comparison for Faster SOC gradient computation}\label{sec:fast_soc_results}

\begin{table}[h!]
    \centering
    \renewcommand{\arraystretch}{1.3} 
    \caption{Results on CIFAR-10}
    \label{table:results_cifar10_timings}
\begin{tabular}{l | ll| l}
\toprule
\textbf{Architecture} &  \multicolumn{2}{|c|}{\textbf{Time per epoch (in seconds)}} &  \textbf{Reduction in}\\
& Ours & Previous & \textbf{Time per epoch}\\
\midrule
LipConvnet-5 & \textbf{25.343} & 30.633 & \textbf{-17.27\%} \\
LipConvnet-10 & \textbf{37.108} & 48.436 & \textbf{-23.39\%} \\
LipConvnet-15 & \textbf{49.274} & 66.741 & \textbf{-26.17\%} \\
LipConvnet-20 & \textbf{61.586} & 83.183 & \textbf{-25.96\%} \\
LipConvnet-25 & \textbf{71.510} & 100.700 & \textbf{-28.99\%} \\
LipConvnet-30 &  \textbf{83.996} & 119.549 & \textbf{-29.74\%} \\
LipConvnet-35 & \textbf{95.065} & 137.100 & \textbf{-30.66\%} \\
LipConvnet-40 & \textbf{106.007} & 156.254 & \textbf{-32.16\%} \\
\bottomrule
\end{tabular}
\end{table}

\begin{table}[h!]
    \centering
    \renewcommand{\arraystretch}{1.3} 
    \caption{Results on CIFAR-100}
    \label{table:results_cifar100_timings}
\begin{tabular}{l | ll| l}
\toprule
\textbf{Architecture} &  \multicolumn{2}{|c|}{\textbf{Time per epoch (in seconds)}} &  \textbf{Reduction in}\\
& Ours & Previous & \textbf{Time per epoch} \\
\midrule
LipConvnet-5 & \textbf{25.034} & 31.877 & \textbf{-21.47\%} \\
LipConvnet-10 & \textbf{36.872} & 49.446 & \textbf{-25.43\%} \\
LipConvnet-15 & \textbf{48.769} & 68.071 & \textbf{-28.36\%} \\
LipConvnet-20 & \textbf{60.416} & 85.007 & \textbf{-28.93\%} \\
LipConvnet-25 & \textbf{71.064} &    101.933 &      \textbf{-30.28\%} \\
LipConvnet-30 & \textbf{86.319} & 122.774 & \textbf{-29.69\%} \\
LipConvnet-35 & \textbf{97.802} & 140.606 & \textbf{-30.44\%} \\
LipConvnet-40 & \textbf{110.345} & 157.503 & \textbf{-29.94\%} \\
\bottomrule
\end{tabular}
\end{table}

\section{Example of projection pooling layer}
\begin{figure}[h!]
\centering
\includegraphics[trim=15cm 7cm 22cm 7cm, clip, width=0.45\linewidth]{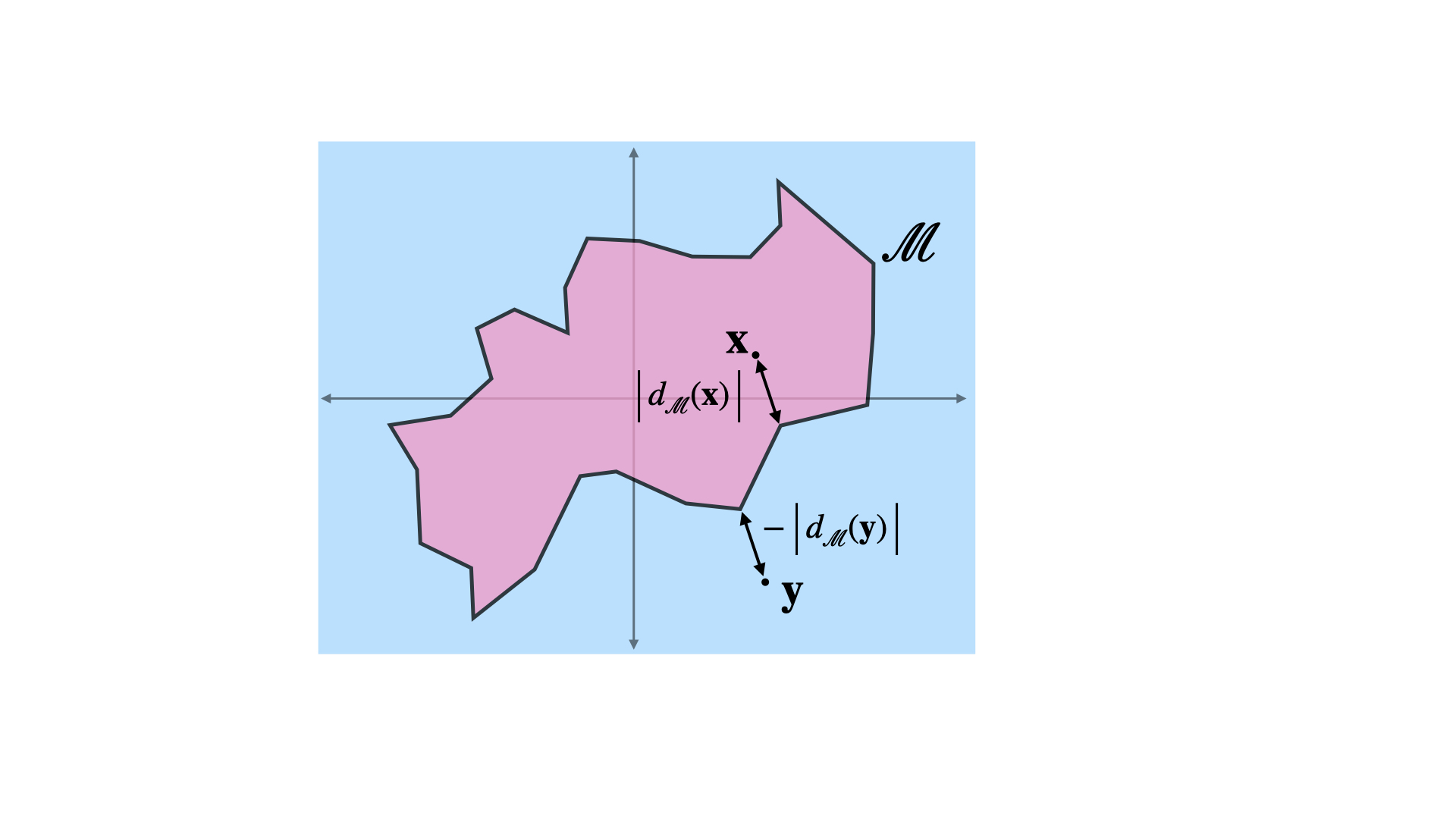}
\caption{Example $2\mathrm{D}$ projection pooling using piecewise linear curve}
\label{fig:lip_pool}
\end{figure}

\end{document}